\documentclass{article}

\usepackage{arxiv}

\usepackage[utf8]{inputenc} 
\usepackage[T1]{fontenc}    
\usepackage{hyperref}       
\usepackage{url}            
\usepackage{booktabs}       
\usepackage{amsfonts}       
\usepackage{nicefrac}       
\usepackage{microtype}      
\usepackage{lipsum}
\usepackage{graphicx}
\usepackage{lineno}
\usepackage{gensymb}
\usepackage{amsmath}
\usepackage{xspace}
\usepackage{arydshln}
\usepackage{amssymb}
\usepackage{bm}
\usepackage{float}

\newcommand{\etal}{\textit{et al.}}

\title{SensorSCAN: Self-Supervised Learning and Deep Clustering for Fault Diagnosis in Chemical Processes}

\author{
  Maksim Golyadkin\\
  AIRI\\
  HSE University\\
  \textit{Moscow, Russia}\\
  \texttt{mgolyadkin@hse.ru}\\
  \And
  Vitaliy Pozdnyakov\\
  AIRI\\
  HSE University\\
  \textit{Moscow, Russia}\\
  \texttt{pozdnyakov@airi.net}\\
  \And
  Leonid Zhukov\\
  HSE University\\
  \textit{Moscow, Russia}\\
  \texttt{lzhukov@hse.ru}\\
  \And
  Ilya Makarov\\
  AIRI\\
  \textit{Moscow, Russia}\\
  \texttt{makarov@airi.net}\\
}

\begin{document}
\maketitle

\begin{abstract}
Modern industrial facilities generate large volumes of raw sensor data during the production process. This data is used to monitor and control the processes and can be analyzed to detect and predict process abnormalities. Typically, the data has to be annotated by experts in order to be used in predictive modeling. However, manual annotation of large amounts of data can be difficult in industrial settings.

In this paper, we propose SensorSCAN, a novel method for unsupervised fault detection and diagnosis, designed for industrial chemical process monitoring. We demonstrate our model's performance on two publicly available datasets of the Tennessee Eastman Process with various faults. The results show that our method significantly outperforms existing approaches (+0.2-0.3 TPR for a fixed FPR) and effectively detects most of the process faults without expert annotation. Moreover, we show that the model fine-tuned on a small fraction of labeled data nearly reaches the performance of a SOTA model trained on the full dataset. We also demonstrate that our method is suitable for real-world applications where the number of faults is not known in advance.  The code is available at \url{https://github.com/AIRI-Institute/sensorscan}.

\end{abstract}

\section{Introduction}

Chemical processing plants use specialized equipment and technology in the manufacturing process. The stability of the production process is usually maintained by a closed-loop control system that can automatically perform small corrections to the operation control parameters to keep the process variables within the desired production range (e.g., reactor temperature, flow velocity, etc.). Process lines are typically well instrumented, with sensors providing live feed to monitoring and control systems. Nevertheless, unexpected process behavior occasionally occurs, both within routine operations (either due to the variability of feedstock or when switching to a new target product) or due to some external factors. This may lead to a decrease in the process yield, to process interruption, and eventually to increased equipment wear or even breakdown.

An important part of a modern process monitoring system is early fault detection and diagnostics. A fault is typically defined as a deviation of a process variable from the acceptable production range \cite{venkatasubramanian_review} -- for example, increased reactor temperature or decreased feed velocity. Identification of faults helps select recovery procedures to return the process to a normal state.

Most modern data-driven fault detection and diagnosis (FDD) methods are developed in a supervised learning setting that requires all sensor data for each time interval to be labeled with the corresponding process state. However, manual labeling of large amounts of data can be expensive and difficult in industrial settings. For example, it may be challenging to determine the exact moment when a process fault starts, due to smoothness of change in the process behavior. In addition, some equipment faults may remain unnoticed due to an early correction during regular technical service. As an alternative, unsupervised FDD methods were proposed for detecting sensor data patterns grouped according to the process states. Traditional unsupervised methods such as principal component analysis (PCA) and Fisher discriminant analysis (FDA) are used to reduce the dimensionality of sensor data, while clustering methods such as k-means and DBSCAN are used to group samples according to process states. The clusters are then manually mapped to fault types so that the trained model can be used as a diagnostic system.

Deep learning methods were proposed to model complex nonlinear relationships between sensors and efficiently process high-dimensional sensor data. Deep learning can be applied to unsupervised FDD by means of feature extraction and subsequent deep clustering, which are alternatives to the traditional dimensionality reduction followed by clustering approach. In recent years, several deep clustering methods have been proposed for unsupervised image classification \cite{cluster5, cluster6, cluster7, cluster8, cluster9}. However, in these works, the feature extractor is randomly initialized and trained in order to improve the quality. As a result, the feature extractor relies on low-level features, thus missing high-level hidden properties of the input data. Self-supervised learning (SSL) methods are aimed at pretraining the feature extractor on pretext tasks to represent the high-level properties of input data \cite{tscp, seqclr, mvts}.

In this paper, we propose a novel unsupervised FDD method, SensorSCAN. The method is based on deep learning techniques designed to achieve high accuracy on chemical sensor data –- namely, SSL and deep clustering. First, a Transformer-based feature extractor \cite{transformer} is pretrained to embed sensor data into a latent space using SSL methods designed for sensor data. Second, a small feedforward network called clustering head is trained to map embeddings onto cluster indices using modified Semantic Clustering by Adopting Nearest Neighbors (SCAN) algorithm \cite{scan}. The clustering training also involves updating feature extraction weights, which makes the latent space separable and leads to faster fault detection. Both techniques (SSL and deep clustering) are combined into a model, which ensures consistency between them.

\begin{figure}[h]
    \centering
    \includegraphics[width=1\linewidth]{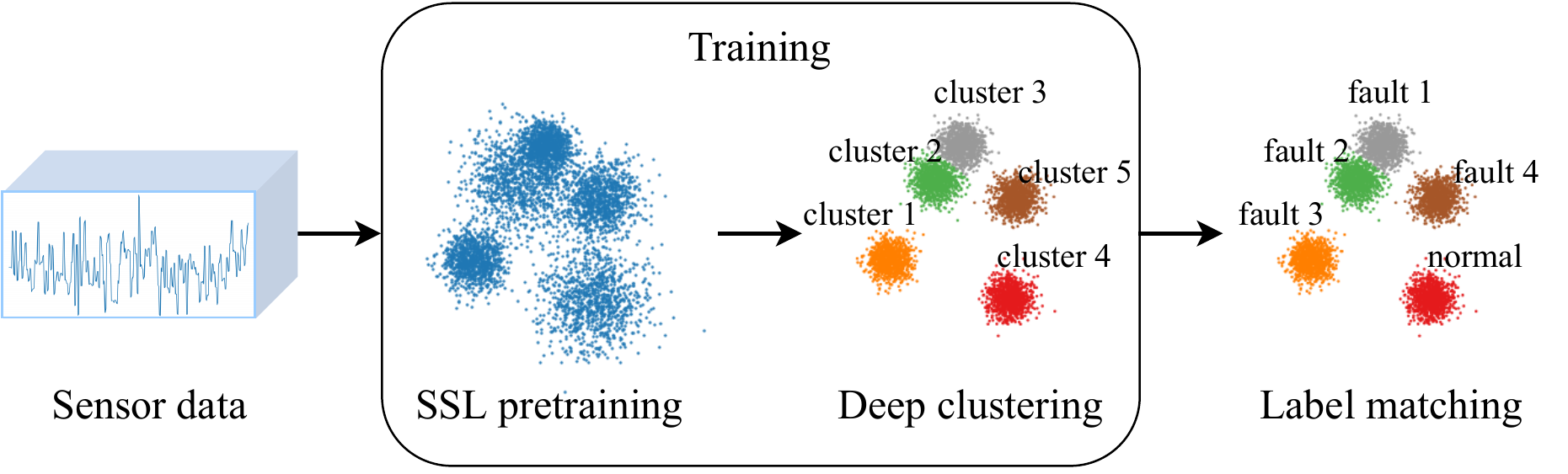}
    \caption{Overview of the SensorSCAN method. First, the feature extractor is trained to map unlabeled sensor data onto latent representations via self-supervised learning. Second, the feature extractor and the clustering head are jointly trained, which makes the latent space well separable. The final step is the manual mapping of the clusters’ indices onto the process states.}
    \label{fig:overview}
\end{figure}

We utilized a label matching technique in our experiments to simulate the manual mapping process performed by experts (see Subsection \ref{label_matching}). In the basic case, a cluster is assigned to a fault that occurs most frequently in that cluster. A schematic view of SensorSCAN is depicted in Figure \ref{fig:overview}. 

Our method demonstrates the best results measured with multiple clustering evaluation metrics on the Tennessee Eastman Process (TEP) benchmarks compared to modern unsupervised FDD methods, such as ConvAE \cite{unsupervised_tep}. Additionally, we implement the semi-supervised setting in which we fine-tune the pretrained model on a small fraction of labeled data. Experimental results show that a model pretrained using SSL methods and then fine-tuned on very small number of labeled examples is close in performance in terms of the True Positive Rate to the SOTA supervised models trained on the full TEP dataset.

The paper is organized as follows. Section \ref{related_works} overviews traditional and state-of-the-art data-driven methods for both supervised and unsupervised FDD, as well as SSL and deep clustering methods designed for sensor data. Section \ref{model} contains a detailed description of the proposed model. Section \ref{tep} presents Tennessee Eastman Process simulators and corresponding datasets. Section \ref{metrics} describes the evaluation metrics and their practical significance. In Section \ref{ablation_study}, we present the results of an ablation study and discuss the sensitivity of the performance to methodological choices, i.e. explore the relationship between the performance and the stability of our model by removing certain components and changing the pretraining setting in order to understand its contribution to the overall model. Section \ref{baselines} describes the models that we consider as the baseline in the experiments. Section \ref{results} overviews the experimental results. In Section \ref{discussion}, we discuss the applicability of our model to real-world industrial settings. Finally, Section \ref{conclusion} contains the summary of the results and outlines directions for future research. 

\section{Related work}
\label{related_works}

Recently, numerous data-driven methods have been proposed for industrial process fault detection and diagnosis \cite{survey1, survey2, more1}. The prevalence and success of data-driven methods are due to the fact that they operate on information extracted from process history, and modern production facilities  generate large amounts of such data. Yet, in most cases, this data is not publicly shared for security reasons, therefore researchers use simulated data \cite{simdata1, simdata2, simdata3} for reproducibility of results and  comparison with existing methods.

A vast majority of data-driven approaches are built utilizing feature extraction from raw historical data. However, data from industrial sensors is highly redundant and intercorrelated, which impedes many data-driven approaches. Feature extraction is beneficial for robustness and improvement of performance since valuable information is often contained in manifolds of lower dimensionality.  The most frequently used feature extraction methods are based on Canonical Correlation Analysis (CCA) \cite{cca1, cca2}, Principal Component Analysis (PCA) \cite{supervised_tep2, unsupervised_tep2, unsupervised_tep3, unsupervised_tep4, pcakmeans}, Partial Least Squares (PLS) \cite{unsupervised_tep5, unsupervised_tep7, unsupervised_tep_pls}, and different variations of autoencoders (AE) \cite{supervised_tep11, supervised_tep5, supervised_tep6, supervised_tep7, supervised_tep8, distillation_process, ensemble_ae, more_unsupervised_1, more_unsupervised_3, more_semisupervised_2}. Over the past decade, deep neural networks have gained popularity in application to industrial processes because they are simultaneously trained to perform feature extraction and classification, which leads to better performances \cite{fdd3, supervised_tep9, supervised_tep1, more2, more3, more4, more5}.

There are two general approaches to fault detection and diagnosis - the supervised and the unsupervised ones. The supervised setting requires a labeled dataset, which means that each time interval must be marked as belonging to a normal or an abnormal process state. As a result, the trained model distinguishes only between the abnormal states observed in the process history. In \cite{cnn_tep}, the authors propose a method based on a deep convolutional neural network. Zhang \etal{} \cite{dbn_tep} show that the Deep Belief Network (DBN) is able to effectively extract features in complex chemical processes. Wang \cite{edbn_tep} presents a modification of the latter architecture called Extended DBN, which alleviates some of the training imperfections. Park \etal{} \cite{lstm_tep} use a combination of autoencoders and LSTM to perform fault detection and fault diagnosis in two stages. Lomov \etal{} \cite{holy_grail} report the Temporal CNN1D2D architecture to simultaneously capture long-running patterns in a single sensor as well as relationships between different sensors.

However, the data which is typically available does not contain all the possible faults; therefore, there is a motivation to use unsupervised methods that are not restricted to a set of previously observed and labeled faults. In \cite{unsupervised_tep_cmeans, unsupervised_tep_svc, unsupervised_tep_pca, unsupervised_tep_gtm, more_unsupervised_2},  alternative approaches to time series clustering and segmentation have been proposed. Application of the CatGAN \cite{catgan} architecture for unsupervised fault diagnosis is reported in \cite{cataae, stft_gen}. The DeepAnT \cite{deepant} architecture utilizes time series prediction for unsupervised anomaly detection. Yu and Yan \cite{unlstm} propose a modified autoencoder architecture with LSTM blocks, called unLSTM, for unsupervised fault detection. Yang \etal{} \cite{gan_tep} and Li \etal{} \cite{madgan} demonstrate models based on discrimination and reconstruction anomaly score of a pretrained GAN. Zheng and Zhao \cite{unsupervised_tep} present a method employing stacked AE, t-SNE, and DBSCAN for clustering and generation of pseudolabels used for supervised training. Rajeevan \etal{} \cite{cognitive} propose an approach that utilizes an incremental one-class neural network for unsupervised anomaly detection and a dynamic shallow neural network for supervised fault diagnosis.

\subsection{Self-supervised learning for time series}

Self-supervised learning methods for time series can be categorized into two groups. Methods from the first group employ pretext tasks that exploit the structural features specific to time series. Oorg et al. \cite{oord} perform time series forecasting in the latent space for representation learning. Another temporal-aware pretext task for self-supervised change point detection is proposed in \cite{tscp}; it consists in determining whether one time series is an extension of another. Approaches from the second group use general representation learning methods which can be applied to any data type and demonstrated the best performance on a variety of machine learning tasks in recent years \cite{GPT-3, coatnet}. Fortuin \etal{} \cite{somvae} develop a modification of variational autoencoder that produces topologically interpretable discrete representations for time series. Mohsenvand et al. \cite{seqclr} apply contrastive learning \cite{SimCLR} to EEG time series. Adaptation of the pretraining routine of a transformer-based BERT \cite{BERT} model to multivariate time series is shown in \cite{mvts}. TS-TCC \cite{tstcc} combines contrastive learning and time series forecasting for representation learning.

\subsection{Deep clustering} 

Basic deep clustering methods are based on application of classical clustering algorithms to the features extracted with a pretrained feature extractor \cite{cluster1, cluster2, cluster3, cluster4}. Typically, feature extractors are autoencoders or models trained with representation learning. These models can be trained with unlabeled data, and their operation is based on extraction of semantically meaningful features from raw data. It is reasonable to expect embeddings of objects to be distributed in the latent space according to the closeness of their semantic meanings rather than raw similarity. However, it often turns out in practice that object embeddings are placed too densely, which complicates the work of the classic clustering algorithms. Alternatively, there exist models trained with end-to-end indirect loss functions to map inputs to cluster indices \cite{cluster5, cluster6, cluster7, cluster8, cluster9}. The key drawback of this approach is that the same loss is used for the feature extraction and the class assignment training processes, i.e. these processes are jointly optimized while having inherently different goals. At the beginning of training, the network is only able to rely on low-level features because the feature extractor is randomly initialized. As learning continues, the model is not able to significantly change the decision-making process. As a result, the network operates only with low-level features ignoring global features, which leads to a suboptimal solution. This shortcoming may be addressed by pretraining the feature extractor and subsequently training the network for cluster assignment with the extracted features. TSUC \cite{tsuc} was the first method employing this idea: the feature extractor is pretrained with a self-supervised learning algorithm, and the classification head is trained with a combination of mutual information loss and contrastive loss. Van Gansbeke \etal{} proposed the SCAN algorithm \cite{scan}, which performs clustering by enforcing the similarity of cluster predictions for neighboring objects with the closeness calculated in the embedding space of the pretrained feature extractor. The results of the arbitrary end-to-end deep clustering algorithm can be improved with the RUC \cite{ruc} add-on module that is based on robust learning techniques.


With respect to existing approaches, our work makes the following contributions:
\begin{itemize}

    \item We introduce a novel unsupervised learning approach designed specifically for the fault detection domain. The approach involves adapting the proven SCAN method to the sensor data previously unexplored in this context. To our knowledge, we are the first to apply deep clustering methods to fault detection.
    \item We propose a new self-supervised learning task based on contrastive learning and masked values reconstruction tailored to the fault detection task by means of appropriate augmentations and mask generation. The original SCAN for images is built upon existing widely available self-supervised pretraining methods. However, there is a need for pretraining methods designed purposely for time series, and we address this gap.
   \item We identify a crucial challenge when applying the SCAN method to the time series domain – namely, that the chunks of time series generated by the sliding window exhibit a higher correlation than images. To mitigate this issue, we develop a subsampling method that effectively handles the high similarities in time series data.
\end{itemize}

\section{Model description}
\label{model}

\begin{figure}
    \centering
    \includegraphics[width=\textwidth]{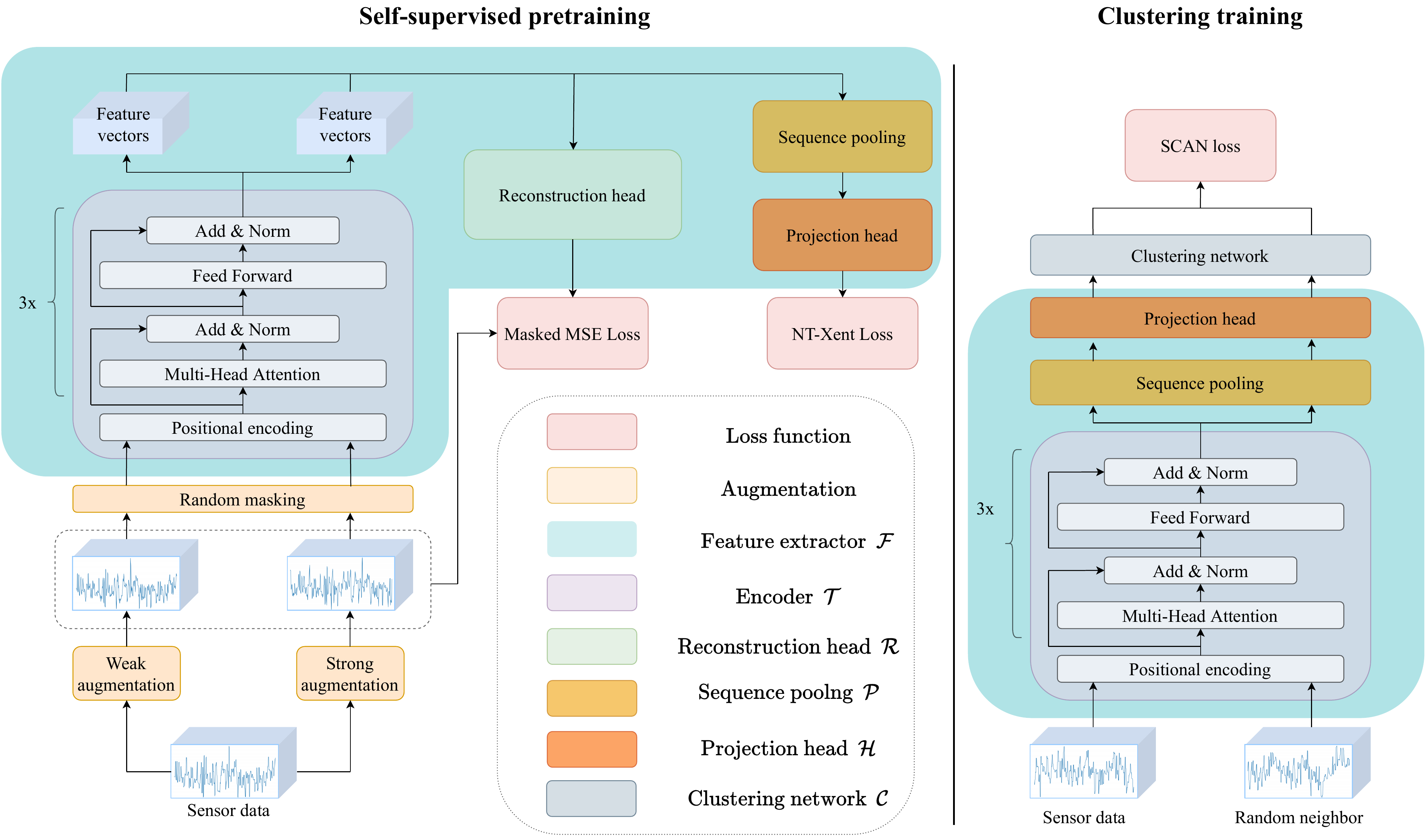}
    \caption{A schematic view of training SensorSCAN. Left: self-supervised pretraining of feature extractor $\mathcal{F}$ with reconstruction and constartive losses. Right: clustering training of feature extractor $\mathcal{F}$ and clustering network $\mathcal{C}$ with SCAN loss.}
    \label{fig:SensorSCAN_scheme}
\end{figure}

The objective of our method is to produce clustering for a given set of unlabeled multivariate time series samples $\mathcal{X} = \{ X_1, \dots, X_N\}$, $X_i \in \mathbb{R}^{L \times D}$, where $L$ is the sample length and $D$ is the number of sensors. The set $X$ has a corresponding set of labels $Y = \{ y_1, \dots, y_N\}$, $y_i \in \{1, \dots, Q\}$ that are not available for training. The produced clusters have to be consistent with the label distribution, which means that samples with the same label should be referred to the same cluster, and samples with different labels should belong to different clusters. This way, the ground truth labels are correctly restored with human assistance. It is also assumed that $Q$ may not be known.

Our model consists of two parts: a large feature extractor $\mathcal{F}$ and a small clustering network $\mathcal{C}$. The feature extractor retrieves characteristic patterns from the data and maps the input samples onto an embedding space of low dimensionality. The clustering network maps an embedding vector onto a vector of  probability distribution over $\tilde{Q}$ clusters, where $\tilde{Q}$ is determined in advance. The training procedure consists of two steps (see Figure \ref{fig:SensorSCAN_scheme}). First, the feature extractor is pretrained with self-supervised learning methods, and the number of clusters is determined via visual analysis of the latent space distribution. Second, the clustering network and the feature extractor are trained using SCAN loss \cite{scan}, and information about the neighbors is obtained from the embedding space. In what follows, we describe each training step in more detail.

\subsection{Self-supervised pretraining}
We use self-supervised learning methods to enable our model to distinguish samples from each other. With these approaches, the neural network explores the internal structure of the data without ground truth labels. The model finds the data-specific patterns containing information about the corresponding process, which are then utilized for clustering. 

We employ the feature extractor based on the Transformer \cite{transformer} architecture. This architecture is highly suitable for sequence processing due to the self-attention modules that provide a global receptive field and the capability to find sophisticated dependencies between the sequence elements.The feature extractor consists of four parts: encoder $\mathcal{T}$, sequential pooling layer $\mathcal{P}$, projection head $\mathcal{H}$, and reconstruction head $\mathcal{R}$.The encoder $\mathcal{T}$ is a three-layer Transformer encoder with the embedding dimensionality $H$. It maps a sample of length $L$ to a sequence of embeddings of the same length, $\mathcal{T} \colon \mathbb{R}^{L \times D} \to \mathbb{R}^{L \times H}$. Sinusoidal positional encoding is used to provides information about the position of each timestamp. The sequential pooling layer $\mathcal{P}$ maps a sequence of the embedding vectors $\tilde{h} = [h_1, \dots, h_L] \in \mathbb{R}^{L \times H}$ onto a single vector $\hat{h} \in \mathbb{R}^{H}$. The pooling is performed with weighted sum, where the weights $w \in \mathbb{R}^{L}$ are obtained with learnable affine transformation and softmax operation:
\begin{equation}
    w = \text{softmax}(W_{pool}^\top\tilde{h}^\top), \: W_{pool} \in \mathbb{R}^{H \times 1},
\end{equation}
\begin{equation}
    \hat{h} = w\tilde{h}.
\end{equation}
The projection head $\mathcal{H}$ is a 2-layer MLP with intermediate BatchNormalization \cite{batchnorm} and ReLU activation. Its goal is to reduce the dimensionality of the encoder's embedding space to encourage the model to encode more informative features and to improve the embedding quality, $\mathcal{H} \colon \mathbb{R}^{H} \to \mathbb{R}^{F}$, where $F$ is the size of the feature extractor embeddings. Finally, the reconstruction head $\mathcal{R}$ is meant to reconstruct raw sensor data from the encoder's embedding vectors, $\mathcal{R} \colon \mathbb{R}^{H} \to \mathbb{R}^{D}$. It is used exclusively for training and consists of one linear layer.

In order to improve the robustness and to achieve higher discriminative power, we employ a combination of two self-supervised learning tasks – masked input reconstruction \cite{mvts} and contrastive learning \cite{SimCLR}. These methods use different procedures to explore data distribution – namely, masked values retrieval and distinguishing between similar and dissimilar samples, respectively. Each of the two methods performs rather well when applied separately, but their pretext tasks effectively complement each other, so we choose to combine them in data preprocessing and loss calculation.

The goal of the first task, mask input reconstruction, is to reconstruct values from the partially masked input. The idea behind this task is that the model has to properly learn the internal structure of the data to recover the values, thus positively affecting the quality of feature extraction and the embeddings. In addition, industrial data may contain missing values due to sensor malfunctions or missing logs and thus, masking adapts our model to such disruptions. For a given sample $X \in \mathbb{R}^{L \times D}$, the binary mask $M \in \{0, 1\}^{L \times D}$ is generated to produce masked samples with element-wise multiplication $\hat{X} = X \odot M$. The mask generation process is based on geometric distribution: for each column corresponding to the measurements of a single sensor, sequences of zeros and ones are consecutively generated, with sequence lengths being independently sampled from geometric distributions with expectations $l_m$ and $l_u$, 
\begin{equation}
\label{mask_length}
    l_u = \frac{1 - r}{r}l_{m},
\end{equation} where $r$ is the ratio of the masked input. Parameters $l_m$ and $r$ determine the complexity of the task and directly affect the characteristics of the obtained embedding space. The geometric distribution is preferred to the Bernoulli distribution because masks generated with the Bernoulli distribution often result in sequences of zeros of unit length. Reconstruction of values from the neighboring points is a relatively simple task in which no informative features are learnt. The loss function for this task is MSE, calculated only for the masked values:
\begin{equation}
    \mathcal{L}_{rec} (\hat{X}, X) = \frac{1}{B} \sum_{i = 1}^{B} \frac{1}{|M^i|} \sum_{l, d \colon M_{ld}^{i} = 0} (\mathcal{R}(\mathcal{T}(\hat{X}))_{ld} - X_{ld})^{2},
\end{equation}
where $B$ is the batch size, and $l$ and $d$ correspond to the number of timestamps and the number of sensors, respectively.

The second task of the self-supervised pretraining is contrastive learning; its core objective is to train the feature extractor so that the embedding vectors of similar samples are closer in the embedding space than  those of dissimilar samples. To achieve this, pairs of similar samples are created using two augmentations applied to the same sample. Next, the model is trained to find mutual patterns within the paired samples and discriminate these samples from the others. Thus, the choice of augmentation type is crucial and depends on the downstream task and the data type. Moreover, augmentation is beneficial in the context of application to industrial data since the amount of data for anomalies is limited and all the possible disturbances are not represented in the historical data.

We use the jitter, scaling, and permutation operations \cite{aug} for data augmentation. Jitter augmentation is an injection of additive Gaussian noise. Scaling augmentation is multiplication by a value sampled from some random distribution, with sampling being independent for each variable in a multivariate time series. Finally, the permutation augmentation consists in splitting a sample into a predetermined number of chunks of random length that are then shuffled and concatenated back into a single sample. Following the approach proposed in \cite{tstcc}, we split augmentations into weak and strong ones to diversify the data so as to improve the model robustness. For the first sample of the pair, we use a combination of scaling and jitter as a weak augmentation, since it modifies the sample slightly. For the second sample, we use a combination of permutation and scaling as a strong augmentation; it breaks time dependencies but preserves the semantic information such as process state.

The training is performed via NT-Xent loss \cite{SimCLR} minimization (normalized temperature-scaled cross entropy). We randomly sample a minibatch of $B$ samples and apply the augmentations to transform it into a minibatch of $2B$ samples. Thus, for each sample there is one positive pair and $(2B - 2)$ negative pairs. The NT-Xent loss makes the distance between positive pairs smaller than the distance between negative pairs. For the positive pair $(i, j)$ it is defined as: 
\begin{equation}
\label{contrastive_loss}
    \mathit{l}_{i, j} = -\log \frac{\exp(\operatorname{sim}(\bm{z_i}, \bm{z_j}) / \tau)}{\sum_{k = 1}^{2B} [k \neq i] \exp(\operatorname{sim}(\bm{z_i}, \bm{z_k}) / \tau)},
\end{equation}
where $\bm{z_i}, \bm{z_j}, \bm{z_k} \in \mathbb{R}^{F}$ denotes the outputs of the projection head, $\operatorname{sim}(\mathbf{u}, \mathbf{v}) = \mathbf{u}^\top\mathbf{v} / (||\mathbf{u}|| \: ||\mathbf{v}||)$ denotes the cosine similarity, and $\tau$ denotes the temperature parameter. The final loss $\mathcal{L}_{cont}$ is calculated across all the $2B$ positive pairs.

The two above tasks affect different latent representations. The first task targets the embeddings generated by the encoder $\mathcal{T}$ for every time series element. The second task influences the embedding of the whole sample, produced from element-wise embeddings with a sequential pooling $\mathcal{P}$ and a projection head $\mathcal{P}$. Thus, there is a potential for combining these methods to improve the efficiency of the pretraining procedure. By reducing the number of errors and improving the quality of the pretraining, we directly affect the final performance; this is crucial because the subsequent clustering training alone would not be able to resolve the fundamental imprecisions of feature extraction.  

Eventually, the SSL pretraining is performed over $E$ epochs. The iteration of pretraining is as follows:
\begin{enumerate}
    \item Randomly sample a minibatch of size B. 
    \begin{equation}
        X = [X_1, \dots, X_{B}]
    \end{equation}
    \item Transform every sample with weak augmentation $\alpha$ and strong augmentation $\beta$.
    \begin{equation}
        \tilde{X} = [\alpha(X_1), \beta(X_1), \dots, \alpha(X_B), \beta(X_B)]
    \end{equation}
    \item Independently for each sample, generate and apply a binary mask $M_i$.
    \begin{equation}
        \hat{X}_i = \tilde{X}_{i} \odot M_i, \: i = \overline{1, 2B}
    \end{equation}
    \item Reconstruct the masked values and calculate the reconstruction score.
    \begin{equation}
        \mathcal{L}_{rec} = \mathcal{L}_{rec}(\hat{X}, \tilde{X})
    \end{equation}
    \item Produce embedding vectors for the masked samples and calculate the NT-Xent loss. 
    \begin{equation}
        \bm{z_i} = \mathcal{H}(\mathcal{P}(\mathcal{T}(\hat{X}_i))), \: i = \overline{1, 2B}
    \end{equation}
    \begin{equation}
        \mathcal{L}_{cont} = \frac{1}{2B} \sum_{k = 1}^{B} (\mathit{l}_{2k-1, 2k} + \mathit{l}_{2k, 2k-1})
    \end{equation}
    \item Calculate the total loss by the weighted sum over the two losses.
    \begin{equation}
    \label{sum_loss}
        \mathcal{L} = \mathcal{L}_{rec} + \lambda_{cont}\mathcal{L}_{cont},
    \end{equation}
    where $\lambda_{cont}$ denotes the weight that adjusts the impact of contrastive loss on joint training.
    \item The feature extractor weights are updated according to the step of the optimization algorithm minimizing loss.
\end{enumerate}

After the training, we discard the reconstruction head, and the feature extractor is defined: $\mathcal{F} = \mathcal{T} \circ \mathcal{P} \circ \mathcal{H}$, $\mathcal{F} \colon \mathbb{R}^{L \times D} \to \mathbb{R}^{F}$; masking and augmentations are no longer applied.

\subsection{Clustering training}

Clustering groups samples according to their similarity, thus separating different process states in the absence of ground truth labels. The SCAN algorithm, initially designed for images, has been shown to exploit the high-quality embeddings of the pretrained feature extractor in the most efficient way, hence we choose it to adapt for our domain.

The training requires preprocessing called nearest neighbors mining. For every sample $X_i$, we retrieve its K nearest neighbors $\mathcal{N}_{X_i}$ in the feature extractor's embedding space. Unlike the original SCAN approach, we do not search for nearest neighbors over the entire training set but over a subsample of the nearest neighbors. The reason is that time series data is usually obtained by applying a sliding window of a small step size. As a result, the nearest neighbors in the embedding space are highly overlapped time series which are not sufficiently representative of the diversity of the data.

The nearest neighbors mining algorithm is as follows:
\begin{enumerate}
    \item Randomly shuffle the training dataset: $\mathcal{X} = [X_1, \dots X_n]$.
    \item Split the dataset into T chunks of equal size: $\mathcal{X} = [\bm{X}_{j_1}, \dots, \bm{X}_{j_T}]$.
    \item For each $X_i$, the nearest neighbors are found within the chunk $\bm{X}_k$ it belongs to:
    $\mathcal{N}_{X_{i}} = \operatorname{NearestNeighbors}(X_i, \bm{X}_k)$.
\end{enumerate}

The clustering network $\mathcal{C}$ is a 2-layer MLP with intermediate BatchNormalization and ReLU activation, and final softmax activation, $\mathcal{C} \colon \mathbb{R}^{F} \to \mathbb{R}^{\tilde{M}}$. The learning process is designed to enforce the same label prediction distribution for both the sample and its neighbors. To avoid the trivial solution when all samples are referred to the same class, the entropy term that penalizes uneven sizes of clusters is used. In total, the loss is computed as follows:

\begin{equation}
    \mathcal{L}_{SCAN} = -\frac{1}{B} \sum_{i = 1}^{B} \log{\langle \mathcal{C}(\mathcal{F}(X_i)),  \mathcal{C}(\mathcal{F}(X_i^{NN}))\rangle} + \lambda_{ent} \: H(\mathcal{C}'),
\end{equation}
\begin{equation}
    \mathcal{C}' = \frac{1}{B} \sum_{i = 1}^{B} \mathcal{C}(\mathcal{F}(X_i)),
\end{equation}
 where $\langle \cdot , \cdot \rangle$ denotes the dot product, $\lambda_{ent}$ denotes entropy loss weight, $X_i^{NN}$ denotes the neighbour randomly sampled from $\mathcal{N}_{X_{i}}$, and $H(\mathcal{C}')$ denotes the entropy over discrete distribution $\mathcal{C}'$.

We would like to notice that end-to-end learning not only provides clustering training but also updates the feature extractor weights, thereby improving disentanglement of its embedding space.

\subsection{Number of clusters} 

In practice, the number of underlying classes is not known upfront, but the SCAN algorithm requires it to be set in advance. As shown in the original paper \cite{scan}, it is possible to employ an overclustering approach by roughly estimating the number of classes. A lower bound on the number of clusters can be obtained by visual inspection of the data after dimensionality reduction with t-SNE \cite{tsne} applied to time series representations  after self-supervised pretraining. 

Since the largest cluster in t-SNE visualization corresponds to the normal behavior, we subsample it to make equally-sized classes for training. The reason is that the entropy term in the loss is sensitive to highly imbalanced data, which is usual for real process history.

In Section \ref{ablation_study}, we analyze the performance of the model with the number of clusters different from the number of classes and advocate for the choice of t-SNE among the other dimensionality reduction approaches, such as PCA \cite{pca} and UMAP \cite{umap}.

The real-world setting requires domain experts to identify faults in the corresponding clusters. It is much easier for an expert to determine that two clusters contain entries with the same fault than to split a cluster containing many faults; therefore, we consider the overclustering approach optimal and practically feasible. 

\subsection{Label matching}
\label{label_matching}

Since one of the fault detection and diagnosis (FDD) goals is to determine the process state, the algorithm is to assign a label to each cluster.

In our experiments, we map cluster index $l$ to a corresponding process state by the weighted maximum occurrence as follows: 
\begin{equation}
\text{LM}(l) = \text{argmax}_{q}\sum_{i=1}^n \alpha_q^l [c_i = l \land y_i = q]
\end{equation}
where $\alpha_q^l$ is the weight of the label $q$ in the cluster $l$. In our experiments, we set $\alpha_q^l = 1$ for all faulty states and $\alpha_q^l = Q_l + 1$ for the normal state, where $Q_l$ is the number of process states in the cluster $l$. In other words, a cluster is assigned to a normal state if the share of normal samples in the cluster is at least $1/(Q_l + 1)$; this parameter’s value helps to reduce false alarms. The label matching procedure is performed only on the training set to obtain the cluster index, i.e. the matching process state that is later used for labeling the test samples. Note that in real-world industrial settings, the cluster-to-label matching has to be performed manually by experts. There exist many labeling approaches that are beyond the scope of the present work, but we assume that labeling just a few samples in each cluster produces the correct matching if the unsupervised model is sufficiently accurate.

\subsection{Fine-tuning on few labeled runs}

Utilization of self-supervised learning techniques makes it possible to efficiently retrieve hidden data distribution without human-made annotations. However, the process history may contain few examples of labeled data that can be used for supervised learning instead of the clustering and label matching approach. In other words, we first perform self-supervised pretraining on unlabeled data. Then, we perform fine-tuning by means of supervised training on the few labeled examples using cross-entropy loss and regularization methods, e.g. label smoothing \cite{labelsmoothing1, labelsmoothing2}. In Section \ref{results}, we show that our model fine-tuned with one simulation run for each fault is able to exhibit results that are almost comparable to the model trained with all data and labels.


\section{Tennessee Eastman Process}
\label{tep}

The Tennessee Eastman Process (TEP) is a well-known benchmark for testing process control and FDD methods. It was created by Eastman Chemical Company and is presented in \cite{original_tep}. The TEP model is based on a real chemical plant process and allows simulating various processes and process faults; a detailed description of TEP can be found in \ref{tep_details}.

Two distinct numerical simulators of the TEP process are available. The first one employs the control scheme from \cite{first_control}. The corresponding dataset (available at \footnote{\url{http://web.mit.edu/braatzgroup/TE_process.zip}, accessed in August 2022}) with 20 faults was presented in \cite{data_simulator}; it consists of a process simulation run for normal operations and for each fault in both the train and the test sets. To increase the diversity of data, an extended version of this dataset \footnote{\url{https://dataverse.harvard.edu/dataset.xhtml?persistentId=doi:10.7910/DVN/6C3JR1}, accessed in August 2022} (hereafter $\text{TEP}_{\text{Rieth}}$) was proposed and employed in \cite{extended_tep}. It was simulated under the same settings as the previous one, but it contains as many as 500 runs with different random seeds for every faulty and normal operation. The second simulator is based on different control schemes \cite{second_control,third_control,fourth_control} and is available from the Tennessee Eastman Challenge Archive \footnote{\url{http://depts.washington.edu/control/LARRY/TE/download.html}, accessed in August 2022}. Primarily, this simulator includes the modified TEP model suggested in \cite{revision}. First, it solves the problem where the result of the simulation depends on the chosen ODE solver if random variation disturbances are activated. Besides, it features 8 additional faults, process measurements, and simulation options. The extended dataset (hereafter $\text{TEP}_{\text{Ricker}}$) generated with this model is presented in \cite{new_ext_tep}.

In $\text{TEP}_{\text{Rieth}}$, there are 500 simulation runs for each process state in both the training and the testing sets, which totals 21,000 simulation runs. A single simulation run consists of 500 and 960 time stamps for training and testing, respectively. The sampling rate is equal to 3 minutes. To make the experimental setup consistent with real fault diagnosis situations, we unbalance the training set by cropping it to 500 runs for the normal state and 5 runs for each faulty state. For testing, we keep the testing set as is.

$\text{TEP}_{\text{Ricker}}$ consists of 2000 time stamps for each simulation run with the sampling rate of 3 minutes. The total run duration is 100 hours, with a fault introduced in the 30-th hour. There are 100 simulation runs for each process state, which totals 2,800 runs. Since there is no train/test split in the original dataset, we split the simulation runs by the ratio of 80/20. Note that this dataset does not have runs that consist only of normal behavior without fault introduction; consequently, we maintain the superiority of the normal state.

\section{Evaluation metrics}
\label{metrics}

We perform FDD on multivariate time series data  $\mathcal{X} = \{ X_1, \dots, X_N\}$, $X_i \in \mathbb{R}^{L \times D}$, where $L$ is the sample length and $D$ is the number of sensors. Each sample is generated by a sliding window of size $L$ with a certain step size. Thereby, each sample is a matrix where every row corresponds to a single time stamp and the column to a sensor. Each sample is assigned with a process state, the set of states being $Y = \{ y_1, \dots, y_N\}$, $y_i \in \{1, \dots, Q\}$. There is one normal state and $Q-1$ faulty states. In our experiments, we set the window size to 100 and the step size to 1; that is, a model has to collect 100 time stamps to make a first prediction. After that, a model is able to predict the process state for every subsequent time stamp. For both datasets, $\text{TEP}_{\text{Rieth}}$ and $\text{TEP}_{\text{Ricker}}$, the window size and the step size correspond to 300 minutes and 3 minutes, respectively. We call  a  sequence of samples a ``run''. Each run starts from the initial process state and continues until the process stops. Runs can be normal or faulty: a normal run contains only normal samples; a faulty run contains several normal samples, while the rest are faulty. That is, we can explicitly define the first faulty sample in each faulty run, which is used in the evaluation of the detection delay. 

\subsection{Clustering metrics} 
Clustering metrics measure the discrepancy between ground truth labels $Y$ and cluster indices $C$ without supevision steps such as label matching.
\begin{itemize}
     \item Unsupervised Clustering Accuracy (ACC) \cite{cai2010locally} is similar to classification accuracy and is calculated as the maximum accuracy over all possible matches between cluster indices and ground truth labels:
    \begin{equation}
        \text{ACC}(Y, C) = \max_f \frac{ \sum_{i = 1}^{n} [y_i = f(c_i)]}{n},
    \end{equation}
    where $n$ is the size of the test dataset and $f$ is a matching function. Calculation of ACC involves solving a maximization problem that can be formulated as the Linear Assignment Problem. The best matching function can be found using the Hungarian Algorithm \cite{kuhn1955hungarian}.
    \item Normalized Mutual Information (NMI) \cite{cai2010locally} is an information-theoretic measure that is equal to the mutual information between ground truth labels and cluster indices normalized by the average of their entropies:
    \begin{equation}
        \text{NMI}(Y, C) = \frac{2I(Y, C)}{H(Y) + H(C)},
    \end{equation}
    where $I(\cdot, \cdot)$ is mutual information and $H(\cdot)$ is entropy. 
    \item Adjusted Rand Index (ARI) \cite{yeung2001details}. The Rand Index (RI) considers all pairs of samples and takes into account the ratio of pairs with the correct cluster index in respect to the ground truth labels:
    \begin{equation}
        \text{RI}(Y, C) = \frac{\sum_{i = 1}^{n} \sum_{\substack{j = 1  \\ i < j}}^{n} [c_i = c_j  \: \land  \: y_i = y_j] + [c_i \neq c_j \: \land \: y_i \neq y_j]}{\binom{n}{2}}
    \end{equation}
    where $[\cdot]$ is the indicator function. The Adjusted Rand Index is the corrected-for-chance version of the Rand index.
\end{itemize}

\subsection{Detection and diagnosis metrics} 

To calculate FDD metrics, we need to compare the ground truth labels and the predicted ones. In the unsupervised setting, each class is linked to some specific cluster. In our experiments, we interconnect classes and clusters using the label matching procedure described in Subsection \ref{label_matching}.

To evaluate the quality of diagnosis, we look at the TPR, FPR, and CDR metrics, where TPR and FPR are calculated separately for each fault, and faulty samples are regarded as positive and normal samples as negative examples: 
\begin{itemize}
    \item TPR$_i$, True Positive Rate, aka Detection Rate -- the number of detected faulty samples of the type $i$ divided by the number of faulty samples of the type $i$.
    \item FPR$_i$, False Positive Rate, aka False Alarm Rate -- the number of false alarms of the type $i$ divided by the number of normal samples. We assume that a model with FPR greater than 0.05 is not applicable in real cases due to inadequately frequent false alarms.
    \item CDR, Correct Diagnosis Rate -- the total number of correctly diagnosed faulty samples divided by the number of detected faulty samples.
\end{itemize}

In addition, we separately measure detection metrics in order to assess the model’s ability to detect whatever faults and then correctly diagnose the type. The detection metrics are: 
\begin{itemize}
\item Detection TPR and Detection FPR -- the TPR and FPR in the binary classification task where all faulty samples count as the positive class and all normal samples as the negative class.
\item ADD, Average Detection Delay -- the average number of samples between the first ground-truth faulty sample and the first detected faulty sample. The averaging is performed across all the faulty runs, excluding the runs with undetected faults (false negatives); the motivation is to keep this metric explicitly interpretable. If the step size is greater than one, then the number of samples is multiplied by the step size. Here we evaluate the delay in the fault detection task, since we consider it important in real-world cases to detect a fault as soon as possible, even if it is misdiagnosed. This allows the operator to prevent accidents by stopping the process or turning on a protection system. 

\end{itemize}

\section{Ablation study and sensitivity to methodological choices}
\label{ablation_study}

In this section, we present an overview of the set of experiments conducted to investigate the sensitivity of our method to the removal of its components, on the one hand, and to methodological choices, on the other. We reviewed the performance of the model under several conditions: by removing the tasks in SSL, by substituting the nearest neighbors mining approach with the original one, and by changing the number of clusters, the training set, and the dimensionality reduction technique. The study was performed on ${\text{TEP}_{\text{Rieth}}}$.

\begin{figure}[!ht]
\begin{minipage}[h]{0.19\linewidth}
\center{\includegraphics[width=1\linewidth]{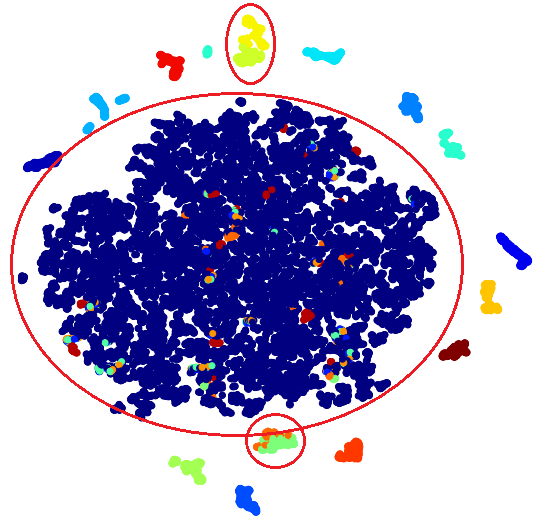}}
\footnotesize
(a)
\end{minipage}
\begin{minipage}[h]{0.19\linewidth}
\center{\includegraphics[width=1\linewidth]{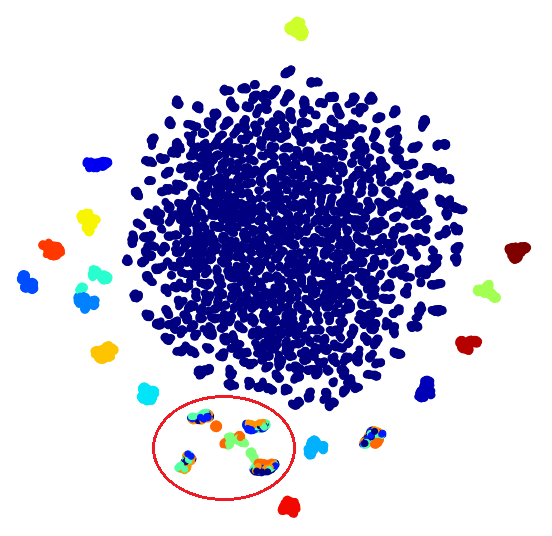}}
\footnotesize
(b)
\end{minipage}
\begin{minipage}[h]{0.19\linewidth}
\center{\includegraphics[width=1\linewidth]{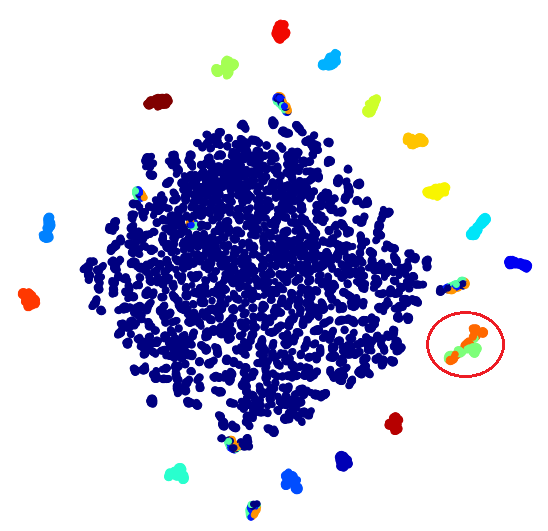}}
\footnotesize
(c)
\end{minipage}
\begin{minipage}[h]{0.19\linewidth}
\center{\includegraphics[width=1\linewidth]{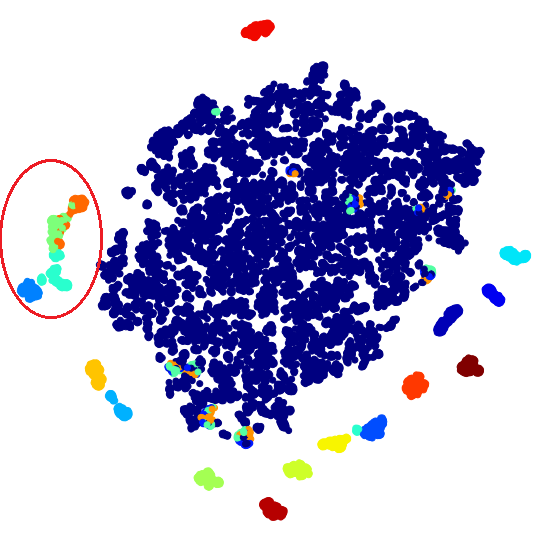}}
\footnotesize
(d)
\end{minipage}
\begin{minipage}[h]{0.19\linewidth}
\center{\includegraphics[width=1\linewidth]{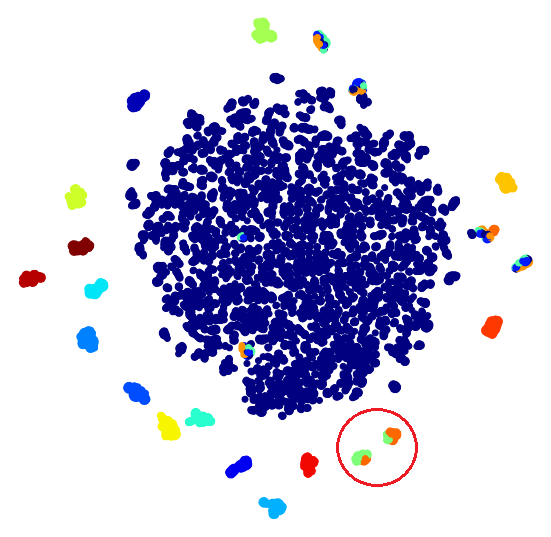}}
\footnotesize
(e)
\end{minipage}
\caption{Visualization of the embedding space with t-SNE on ${\text{TEP}_{\text{Rieth}}}$. The colors correspond to the ground truth labels. Left to right: reconstruction task (a), contrastive learning (b), both tasks (c), both tasks with naive SCAN (d), ours (e). Our model produces denser clusters and decreases the size of the large mixed group. The red ellipses highlight the key differences between the figures.}
\label{fig:ablation}
\end{figure}

We carried out the pretraining with masked input reconstruction and the contrastive learning tasks separately to show that they complement each other; we also compared the performance of our nearest neighbors mining approach to the original one (proposed in \cite{scan}) for the model pretrained with both of the self-supervised tasks. The visual results of the study are depicted in Figure \ref{fig:ablation}, and the numerical outcomes are presented in Table \ref{tab:ablation}. We used the clustering method proposed in \cite{unsupervised_tep} for the first three configurations.

It may be noticed that the combination of self-supervised methods significantly improves disentanglement of the embedding space and increases the number of discriminated classes (Figure \ref{fig:ablation}(c)). It is also noticeable that SCAN with nearest neighbors mining (Figure \ref{fig:ablation}(d)) proposed in the original paper yields worse results than the method proposed in \cite{unsupervised_tep}. However, SCAN with our modification of nearest neighbors mining outperforms the latter. Substantial improvement can be seen in Figure \ref{fig:ablation}(e), when Fault 10 and Fault 16 (the orange and the green dots) are separated into two clusters.

\begin{figure}[h]
    \centering
    \includegraphics[width=0.5\linewidth]{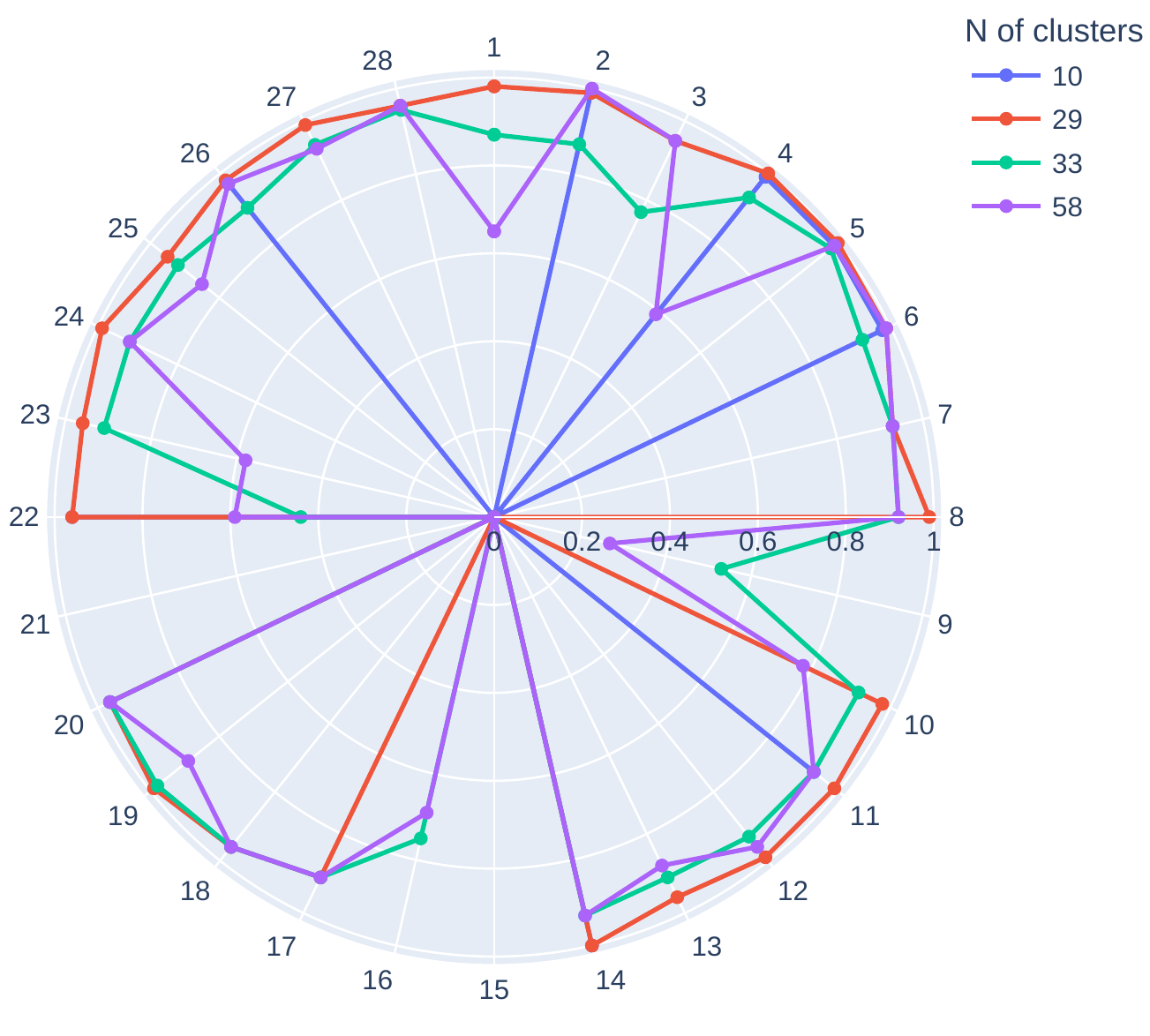}
    \caption{Radar chart with TPR values evaluated on ${\text{TEP}_{\text{Ricker}}}$ for various number of clusters. Faults are numbered on the circle; the distance between 0 and the points represents TPR values; the points in 0 represent undetected faults. FPR values are omitted from the chart since they all are below the $0.05$ threshold.}
    \label{fig:polar_n_clusters}
\end{figure}

We also conducted a series of experiments to evaluate how the predetermined number of clusters $\tilde{M}$ affects the final performance;  $\text{TEP}_{\text{Ricker}}$ was used since it contains a higher number of faults. In Figure \ref{fig:polar_n_clusters} and Table \ref{tab:benchmark2_clustersize_agg}, we show that overclustering does not drop the performance significantly (the FPR values are omitted from the chart since they all are below the $0.05$ threshold; the exact TPR and FPR values can be found in \ref{appendix_metrics}). When using twice as many clusters as the classes, the result is almost identical to using the correct number of clusters — however, twice as much human labor is required for cluster labeling. With a small increase in the number of clusters, the performance of the model even improves where the need for human labor remains essentially the same. Imperfect pretraining produces an embedding space with blobs containing samples of two or more classes; however, overclustering allows separating such blobs into pure clusters. In contrast, underclustering significantly reduces performance due to the fact that at least one cluster will contain multiple ground truth classes.

\begin{figure}[h]
    \centering
    \includegraphics[width=0.5\linewidth]{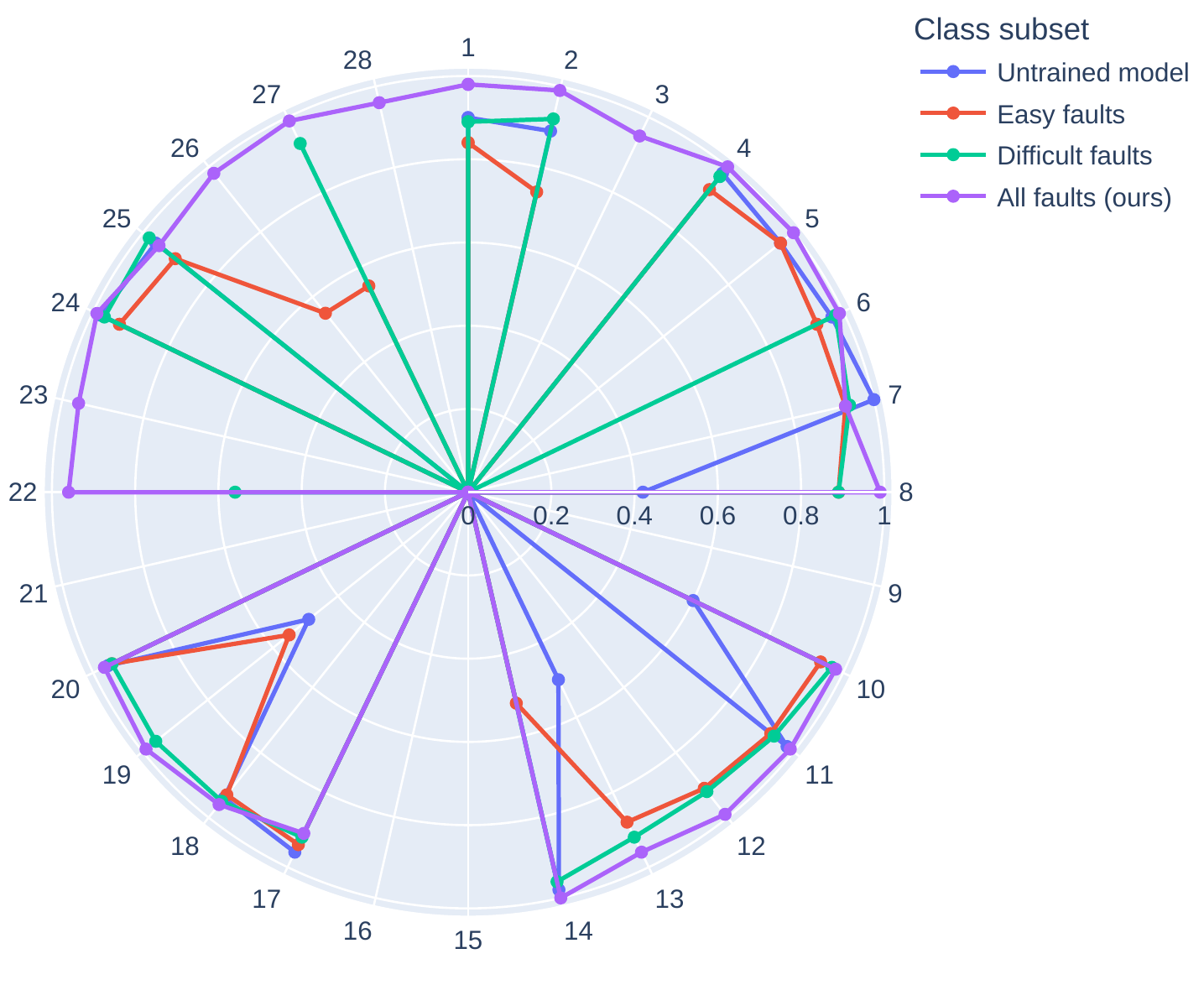}
    \caption{Radar chart with TPR values evaluated on ${\text{TEP}_{\text{Ricker}}}$ for various faults. Faults are numbered on the circle, the distance between 0 and the points represents TPR values; the points in 0 represent undetected faults. FPR values are omitted from the chart since they all are below the $0.05$ threshold.}
    \label{fig:polar_fault_type}
\end{figure}

To demonstrate the high generalizability of the pretrained feature extractor, we compared the performance of the models trained in the first step with different subsets of faults; the results are reported in Figure \ref{fig:polar_fault_type} and Table \ref{tab:benchmark2_faults_ablation_agg} (the FPR values are omitted from the chart since they all are below the 0.05 threshold; the exact TPR and FPR values are found in \ref{appendix_metrics}). Faults were divided into two equal subsets based on their difficulty, as shown in Table \ref{tab:benchmark2_unsupervised}: Faults 4, 5, 6, 7, 8, 11, 12, 14, 19, 20, 23, 24, and 27 were regarded as easy, and the rest were considered difficult. In the second training step, all the faults were used. We also used an untrained model with randomly initialized weights for comparison. The results show that the majority of faults that were not previously seen by the feature extractor are separable in its latent space. However, the presence of a fault in the pretraining dataset does not guarantee its detection. In addition, training can even deteriorate the quality of the embeddings for the samples of this fault (see Fault 26 from the "Difficult faults" model), which is clearly seen in comparison with the untrained model. However, training increases the CDR for detected errors and improves fault diagnosis, especially when training with the difficult faults. To summarize, our model benefits from fault diversity that facilitates productive pretraining and results in high retrieval capability.

\begin{figure}[!ht]
    \begin{minipage}[h]{0.32\linewidth}
    \center{\includegraphics[width=1\linewidth]{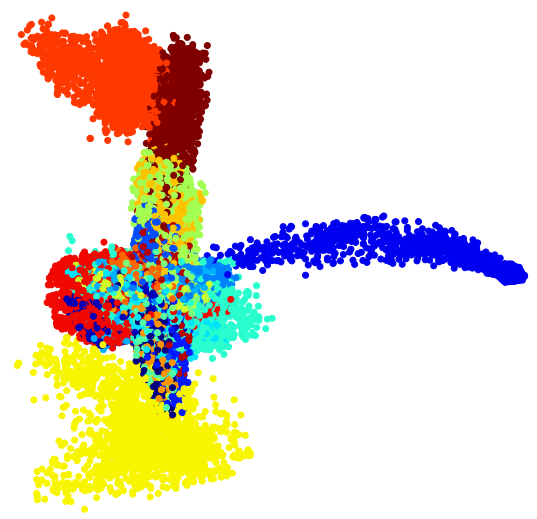}}
    \end{minipage}
    \begin{minipage}[h]{0.32\linewidth}
    \center{\includegraphics[width=1\linewidth]{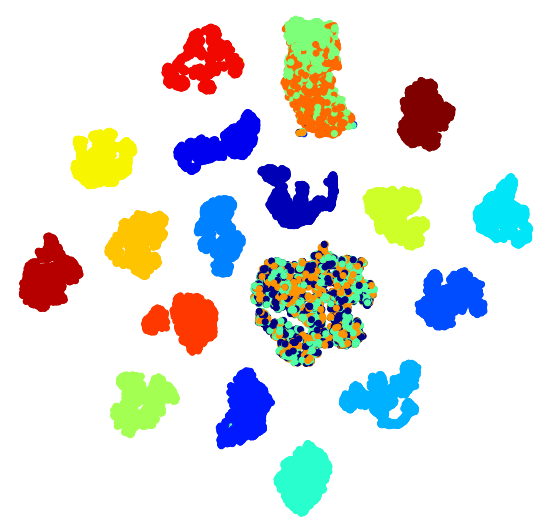}}
    \end{minipage}
    \begin{minipage}[h]{0.32\linewidth}
    \center{\includegraphics[width=1\linewidth]{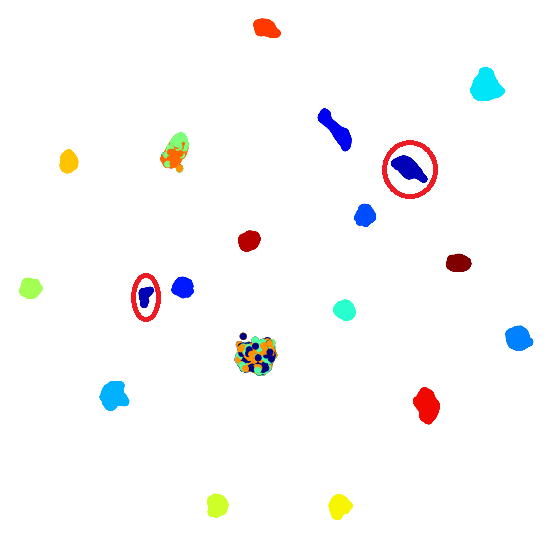}}
    \end{minipage}
    \caption{Comparison of dimensionality reduction algorithms on $\text{TEP}_{\text{Rieth}}$ (here, we use the full dataset to improve the quality of visualisation and to demonstrate the reasonableness of our choice). The colors correspond to the ground truth labels. Left to right: PCA, t-SNE, UMAP. PCA failed to preserve the data distribution, while UMAP has spread the samples of the same process state far apart (as indicated with red circles).}
    \label{fig:reduction_comp1}
\end{figure}

Finally, we carried out a visual comparison to justify the choice of the dimensionality reduction algorithm (see Figure \ref{fig:reduction_comp1}). PCA fails to reduce the dimensionality to two dimensions whilst preserving the data distribution. On the opposite, t-SNE and UMAP adequately represent the data division into groups according to the ground truth labels, which are unknown to the algorithms. Examination of hyperparameters found that UMAP divides samples of the same process state into several groups more frequently than t-SNE, so we decided to utilize the latter in our experiments.

\begin{table}[!ht]
    \centering
    \caption{Aggregated detection and diagnosis metrics evaluated on ${\text{TEP}_{\text{Ricker}}}$ for various numbers of clusters.}
    \label{tab:benchmark2_clustersize_agg}
    \begin{tabular}{lcccc}
    \hline
         &  10 & 29 & 33 & 58 \\
    \hline
    Detection TPR & $\phantom{0}$0.82  & $\phantom{0}$0.87 & $\phantom{0}$\textbf{0.89} & $\phantom{0}$0.86\\
    Detection FPR & $\phantom{0}$0.00 & $\phantom{0}$0.00 & $\phantom{0}$0.01 & $\phantom{0}$0.01\\
    CDR  & $\phantom{0}$0.34 & $\phantom{0}$\textbf{0.96} & $\phantom{0}$0.92 & $\phantom{0}$0.91\\
    ADD  & 54.08 & \textbf{28.47} & 45.50 & 61.91\\
    \hline
    \end{tabular}
\end{table}

\begin{table}[!ht]
    \centering
    \caption{Aggregated detection and diagnosis metrics evaluated on ${\text{TEP}_{\text{Ricker}}}$ for various faults.}
    \label{tab:benchmark2_faults_ablation_agg}
    \begin{tabular}{p{25mm}cccc}
    \hline
         & Untrained model & Easy faults & Difficult faults & All faults (ours) \\
    \hline
    Detection TPR & $\phantom{0}$0.62 & $\phantom{00}$0.68 & $\phantom{00}$0.66 & \textbf{$\phantom{0}$0.87} \\
    Detection FPR & $\phantom{0}$0.00 & $\phantom{00}$0.01 & $\phantom{00}$0.00 & $\phantom{0}$0.00 \\
    CDR   & $\phantom{0}$0.83 & $\phantom{00}$0.87 & $\phantom{00}$0.95 & \textbf{$\phantom{0}$0.96} \\
    ADD & 67.36 & 150.14 & 124.86 & \textbf{28.47} \\
    \hline
    \end{tabular}
\end{table}

\begin{table}[!ht]
    \centering
    \caption{Results of ablation study. Clustering metrics evaluated on $\text{TEP}_{\text{Rieth}}$.}
    \label{tab:ablation}
    \begin{tabular}{lccc}
    \hline
         &  ACC & ARI  & NMI\\
    \hline
    Only reconstruction task    & 0.632 & 0.546 & 0.711\\
    Only contrastive learning    & 0.730 & 0.531 & 0.804\\
    Both tasks    & 0.780 & 0.697 & 0.838\\
    Both tasks with naive SCAN    & 0.756 & 0.659 & 0.812\\
    Ours            & \textbf{0.785} & \textbf{0.703} & \textbf{0.846}\\
    \hline
    \end{tabular}
\end{table}

\begin{figure}[!ht]
\begin{minipage}[h]{0.24\linewidth}
\center{\includegraphics[width=1\linewidth]{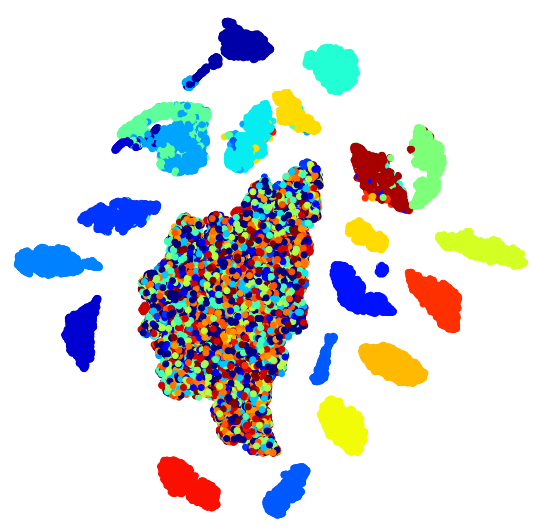}}
\end{minipage}
\begin{minipage}[h]{0.24\linewidth}
\center{\includegraphics[width=1\linewidth]{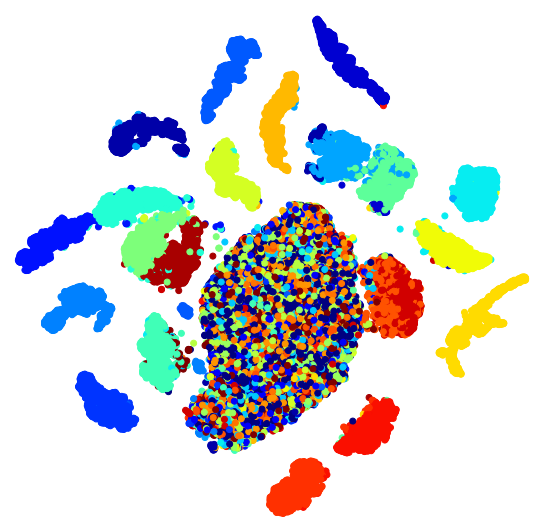}}
\end{minipage}
\begin{minipage}[h]{0.24\linewidth}
\center{\includegraphics[width=1\linewidth]{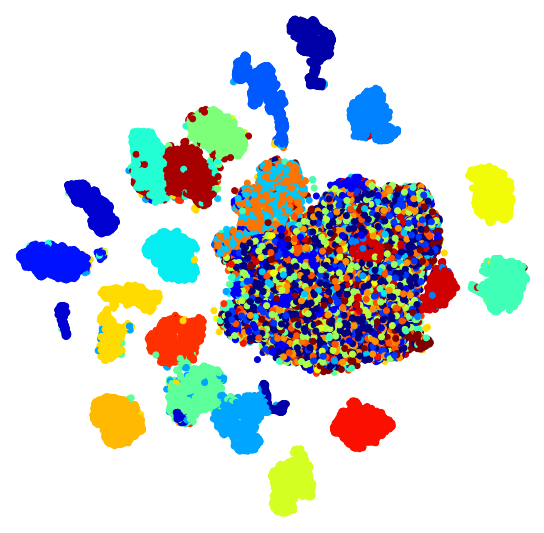}}
\end{minipage}
\begin{minipage}[h]{0.24\linewidth}
\center{\includegraphics[width=1\linewidth]{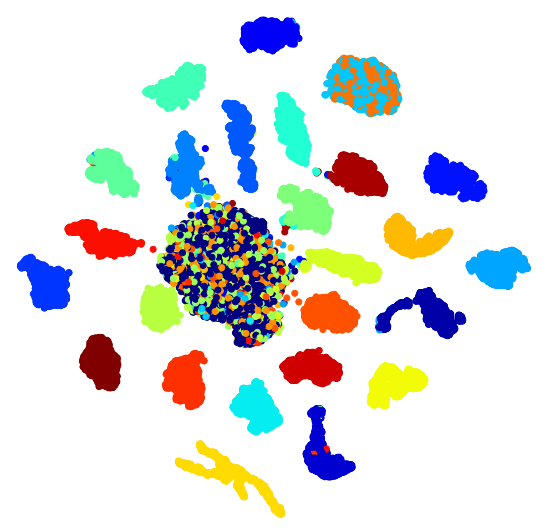}}
\end{minipage}
\caption{Comparison of feature extractor pretraining with various faults for ${\text{TEP}_{\text{Ricker}}}$. The colors correspond to the ground truth labels. Left to right: untrained model, easy faults, difficult faults, all faults (ours). Pretraining on all faults decreases the size of the large mixed group.}
\label{fig:ablation_faults_ricker}
\end{figure}


\section{Baselines}
\label{baselines}

We compared SensorSCAN with unsupervised models of different types to cover both traditional and latest data-driven methods. We assume that an accurate unsupervised model can detect all faults by dividing samples into separate groups, so we set the number of clusters equal to the number of classes. For more detail on the training setup and the hyperparameter tuning for SensorSCAN, see \ref{appendix_training}.

\textbf{PCA.} We consider PCA with k-means clustering as a simple model with cheap computational costs; the approach was proposed in \cite{pcakmeans}. We perform PCA to represent the samples in a 25-dimensional embedding space and then apply k-means to determine the clusters of samples. The dimensionality of the embedding space is selected according to the largest TPR on the training set.

\textbf{ST-CatGAN.} ST-CatGAN is a deep learning FDD method based on a convolutional neural network with adversarial training. The values of hyperparameters were taken from \cite{stft_gen} and adapted to the TEP datasets. Samples of the shape $(100, D)$ where $D$ is the number of sensors, are converted with the short-time Fourier transformation (STFT)  into multi-channel matrices of the shape $(16, 16, D)$. The window size of the STFT is 30, the step size is 23. The values of the hyperparameters are chosen according to the largest TPR values on the training set.

\textbf{ConvAE.} The approach consists of three steps: feature extraction by convolutional stacked autoencoders, converting the obtained embeddings into 2-dimensional vectors with the t-SNE algorithm, and performing k-means to determine the clusters of samples. The method was proposed in \cite{unsupervised_tep} as a state-of-the-art unsupervised FDD for TEP; we borrow the hyperparameter values from the paper.

We also consider the following supervised model to compare with fine-tuned SensorSCAN.

\textbf{GRU.} The Gated Recurrent Unit (GRU) was adapted to FDD in \cite{holy_grail}. Following the paper, we regard the model as a state-of-the-art supervised FDD for TEP. The model we use is referred to as ”GRU type:2”, with the values of hyperparameters proposed in the paper.

\section{Experimental results}
\label{results}

\subsection{Unsupervised setting} 

The unsupervised setting consists of two steps – namely, clustering of samples which is followed by label matching. The clustering metrics were calculated based on cluster indices without the use of label  matching; the results are reported in Table \ref{tab:clustering1} and Table \ref{tab:clustering2} for $\text{TEP}_{\text{Rieth}}$ and $\text{TEP}_{\text{Ricker}}$, respectively. PCA and ST-CatGAN showed the worst results compared to their competitors across all the clustering metrics, while SensorSCAN yielded the best results. 

\begin{table}[!ht]
    \centering
    \caption{Clustering metrics evaluated on ${\text{TEP}_{\text{Rieth}}}$.}
    \label{tab:clustering1}
    \begin{tabular}{lccc}
    \hline
         &  ACC & ARI  & NMI\\
    \hline
    PCA             & 0.274 & 0.110 & 0.363\\
    ST-CatGAN       & 0.175 & 0.113 & 0.222\\
    ConvAE          & 0.402 & 0.124 & 0.467\\
    Ours            & \textbf{0.785} & \textbf{0.703} & \textbf{0.846}\\
    \hline
    \end{tabular}
\end{table}

\begin{table}[!ht]
    \centering
    \caption{Clustering metrics evaluated on ${\text{TEP}_{\text{Ricker}}}$.}
    \label{tab:clustering2}
    \begin{tabular}{lccc}
    \hline
         &  ACC & ARI  & NMI\\
    \hline
    PCA             & 0.352 & 0.132 & 0.448\\
    ST-CatGAN       & 0.302 & 0.129 & 0.361\\
    ConvAE          & 0.523 & 0.239 & 0.573\\
    Ours            & \textbf{0.736} & \textbf{0.481} & \textbf{0.850}\\
    \hline
    \end{tabular}
\end{table}

\begin{figure}[h]
    \centering
    \includegraphics[width=1.0\linewidth]{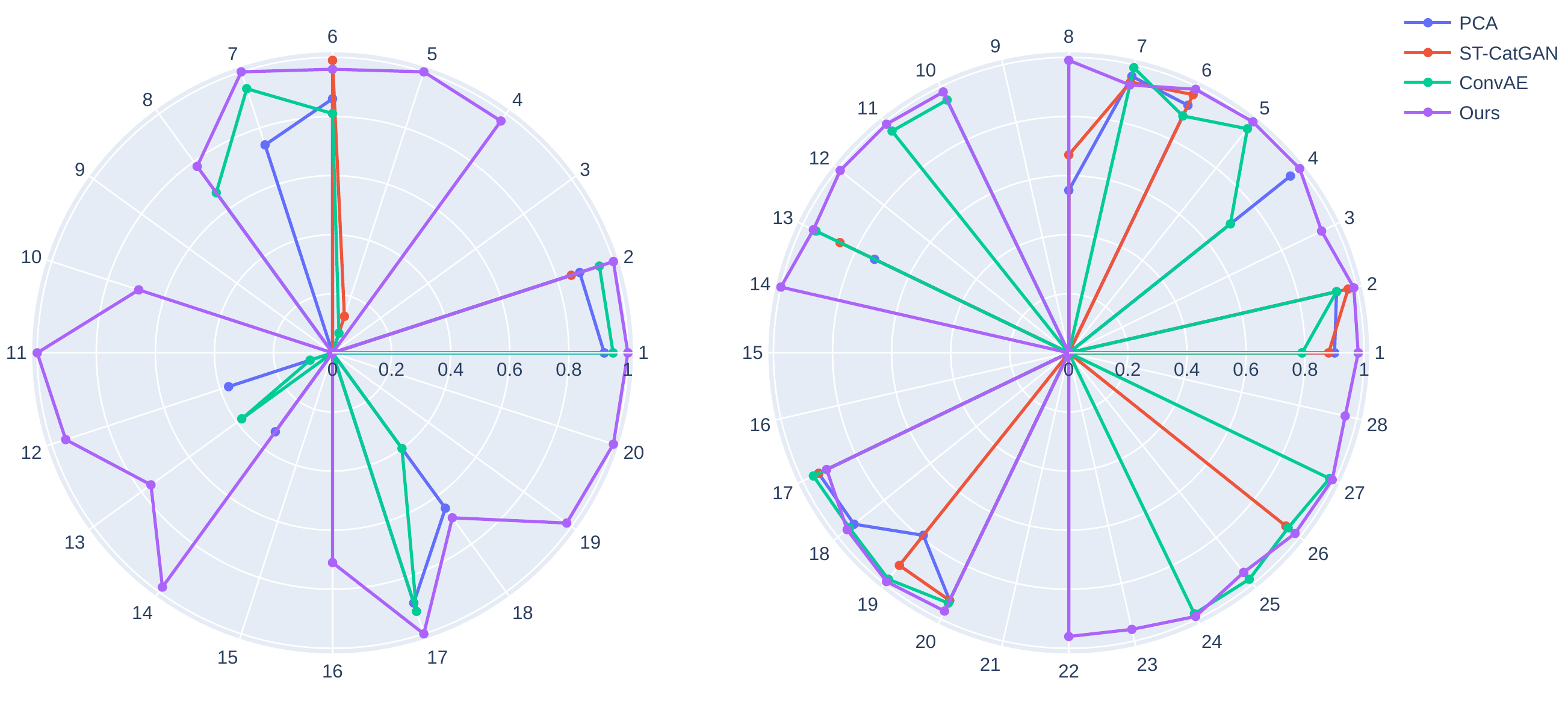}
    \caption{Radar chart with TPR values evaluated on ${\text{TEP}_{\text{Rieth}}}$ (left) and ${\text{TEP}_{\text{Ricker}}}$ (right) in the unsupervised setting. Faults are enumerated on the circle, the distance between 0 and the points represents TPR values, the points in 0 represent undetected faults.}
    \label{fig:polar_fdd}
\end{figure}

The FDD metrics are calculated using the label matching output – see Figure  \ref{fig:polar_fdd} and Tables \ref{tab:benchmark1_unsupervised_agg}, \ref{tab:benchmark2_unsupervised_agg}. The FPR values are omitted from the chart since they all are below the $0.05$ threshold (the exact TPR and FPR values can be found in \ref{appendix_metrics}). We observe that PCA is able to effectively classify the normal type to prevent false alarms, but at the same time the majority of faults (12 out of 20 and 17 out of 28, respectively) remain undetected. ConvAE significantly outperformed PCA: it is able to detect all the faults that can be detected by PCA, including faults 5, 7, 14 in $\text{TEP}_{\text{Rieth}}$ and faults 5, 10, 11, 24, 25, 26, 27 in $\text{TEP}_{\text{Ricker}}$ (the description of faults can be found in Table \ref{tab:faults}). The majoriy of  the detected faults has high TPR values. However, there still remain undetected faults (8 out of 20 and 11 out of 28, correspondingly). SensorSCAN is able to detect almost all the faults except faults 9 and 15 in $\text{TEP}_{\text{Rieth}}$ and faults 9, 15, 16, and 21 in $\text{TEP}_{\text{Ricker}}$. Note that these faults remain undetected by all the unsupervised models, so we term them hard-to-detect faults; in the next subsection, we show that SensorSCAN is able to detect some of them with the help of fine-tuning. All the models are operating with almost zero Detection FPR; thus, they can be applied in real cases since they do not cause excessive numbers of false alarms. SensorSCAN showed the best results in the aggregated detection and diagnosis metrics, Detection TPR and Detection  CDR, detecting 90\% of faults in $\text{TEP}_{\text{Rieth}}$ and 87\% in $\text{TEP}_{\text{Ricker}}$. The detected faults were correctly diagnosed almost always (96\% on both datasets). The average detection delay was significantly shorter with respect to the other models, amounting to 27.7 and 28.47 time stamps, respectively. The next fastest model was ConvAE, which detects faults within about 49.95 and 52.28 time stamps, respectively. 

\begin{table}[!ht]
    \centering
    \caption{Aggregated detection and diagnosis metrics evaluated on ${\text{TEP}_{\text{Rieth}}}$ in the unsupervised setting.}
    \label{tab:benchmark1_unsupervised_agg}
    \begin{tabular}{p{25mm}cccc}
    \hline
         &  PCA & ST-CatGAN  & ConvAE & Ours\\
    \hline
    Detection TPR & $\phantom{0}$0.36 & $\phantom{00}$0.30 & $\phantom{0}$0.48 & \textbf{$\phantom{0}$0.84}\\
    Detection FPR & $\phantom{0}$0.00 & $\phantom{00}$0.00 & $\phantom{0}$0.00 & $\phantom{0}$0.00\\
    CDR   & $\phantom{0}$0.79 & $\phantom{00}$0.32 & $\phantom{0}$0.93 & \textbf{$\phantom{0}$0.92}\\
    ADD & 113.95 & 102.63 & 49.95 & \textbf{5.21}\\
    \hline
    \end{tabular}
\end{table}

\begin{table}[!ht]
    \centering
    \caption{Aggregated detection and diagnosis metrics evaluated on ${\text{TEP}_{\text{Ricker}}}$ in the unsupervised setting.}
    \label{tab:benchmark2_unsupervised_agg}
    \begin{tabular}{p{25mm}cccc}
    \hline
         &  PCA & ST-CatGAN  & ConvAE & Ours \\
    \hline
    Detection TPR & $\phantom{00}$0.36 & $\phantom{00}$0.36 & $\phantom{0}$0.64 & \textbf{$\phantom{0}$0.87} \\
    Detection FPR & $\phantom{00}$0.00 & $\phantom{00}$0.00 & $\phantom{0}$0.00 & $\phantom{0}$0.00 \\
    CDR   & $\phantom{00}$0.95 & $\phantom{00}$0.89 & $\phantom{0}$0.89 & \textbf{$\phantom{0}$0.96} \\
    ADD & 111.49 & 135.04 & 52.28 & \textbf{28.47} \\
    \hline
    \end{tabular}
\end{table}

\begin{figure}[h]
    \begin{minipage}[h]{0.24\linewidth}
    \center{\includegraphics[width=1\linewidth]{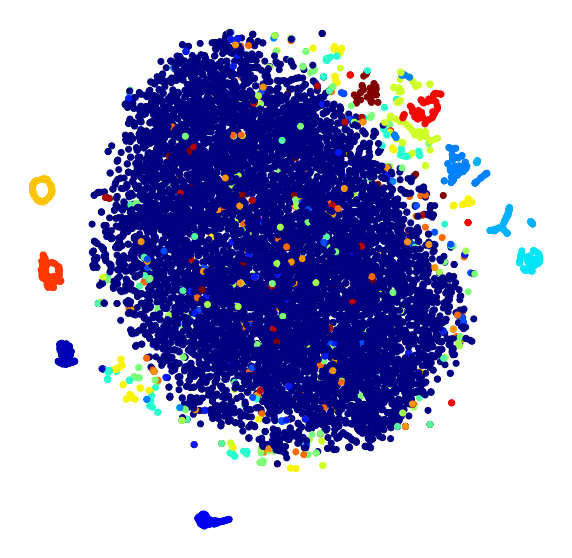}}
    \end{minipage}
    \begin{minipage}[h]{0.24\linewidth}
    \center{\includegraphics[width=1\linewidth]{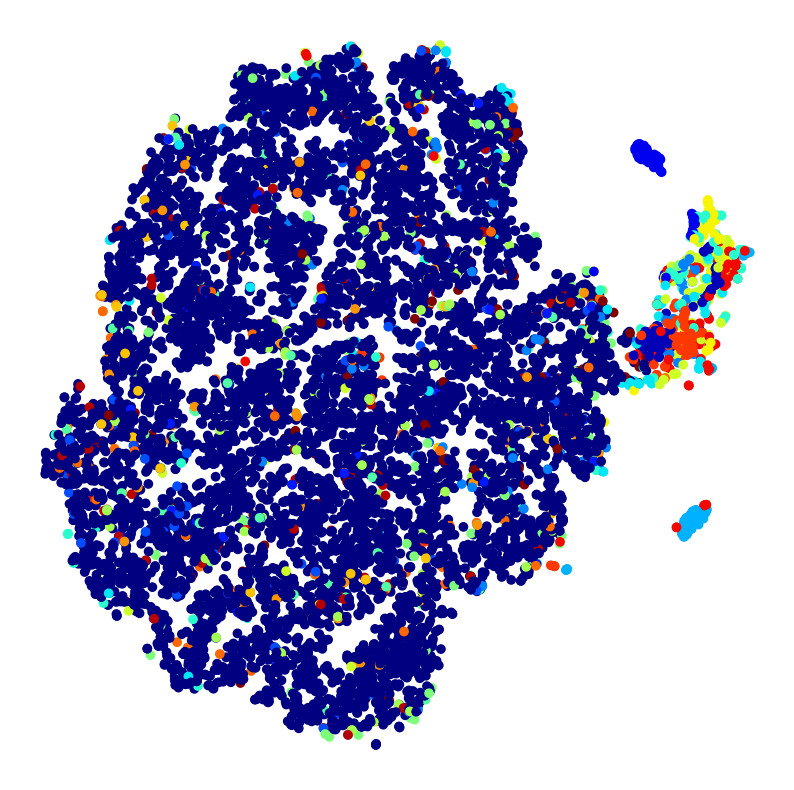}}
    \end{minipage}
    \begin{minipage}[h]{0.24\linewidth}
    \center{\includegraphics[width=1\linewidth]{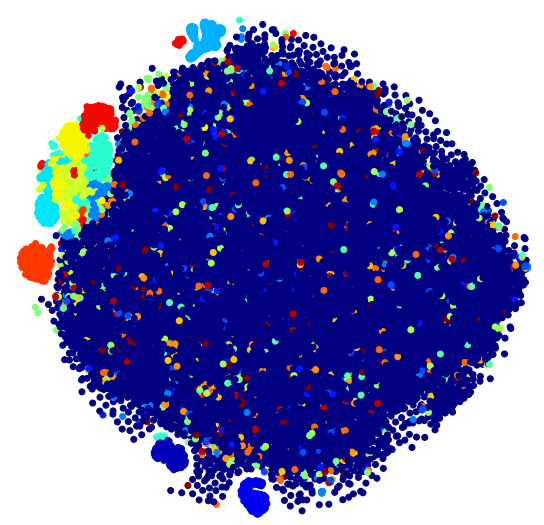}}
    \end{minipage}
    \begin{minipage}[h]{0.24\linewidth}
    \center{\includegraphics[width=1\linewidth]{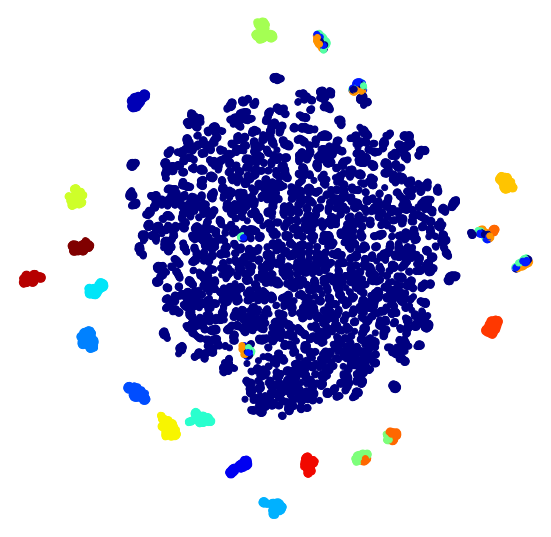}}
    \end{minipage}
    \caption{Model comparison on $\text{TEP}_{\text{Rieth}}$; visualization of the embedding space with t-SNE. The colors correspond to the ground truth labels. Left to right: PCA, ST-CatGAN, ConvAE, ours. Our model is able to separate most faults from the normal state cluster (blue blob), while the other models fail to separate most of the faults from the normal state and from each other.}
    \label{fig:tsne_results1}
\end{figure}

\begin{figure}[!ht]
    \begin{minipage}[h]{0.24\linewidth}
    \center{\includegraphics[width=1\linewidth]{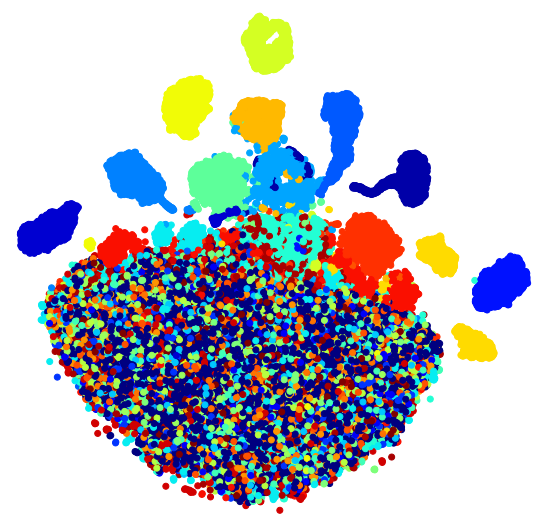}}
    \end{minipage}
    \begin{minipage}[h]{0.24\linewidth}
    \center{\includegraphics[width=1\linewidth]{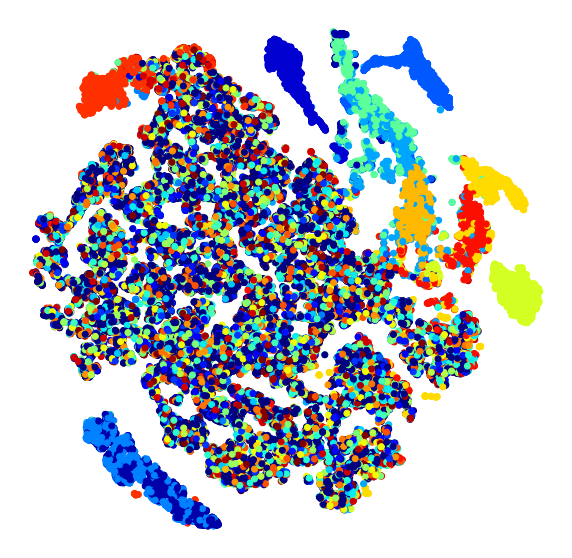}}
    \end{minipage}
    \begin{minipage}[h]{0.24\linewidth}
    \center{\includegraphics[width=1\linewidth]{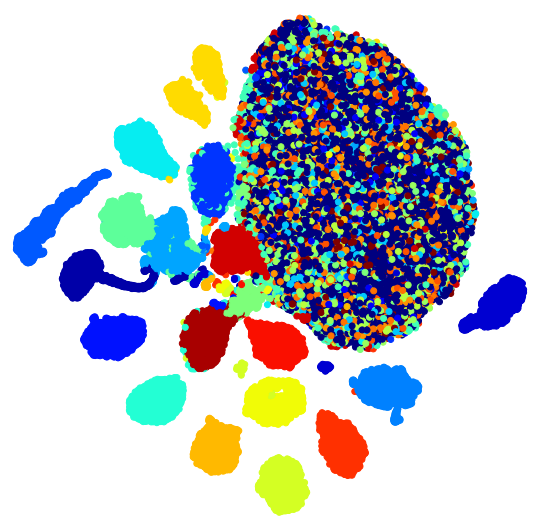}}
    \end{minipage}
    \begin{minipage}[h]{0.24\linewidth}
    \center{\includegraphics[width=1\linewidth]{figures/our_all_faults_ricker.png}}
    \end{minipage}
    \caption{Model comparison on $\text{TEP}_{\text{Ricker}}$; visualization of the embedding space with t-SNE. The colors correspond to the ground truth labels. Left to right: PCA, ST-CatGAN, ConvAE, ours. We can see that our model produces denser clusters as well as decreases the size of the large mixed group.}
    \label{fig:tsne_results2}
\end{figure}

In order to evaluate the ability of the unsupervised models to diagnose faults, we conduct a 2-dimensional t-SNE visualization where the dots represent embeddings and are colored in accordance with the ground truth labels. First, let us consider $\text{TEP}_{\text{Rieth}}$ in Figure \ref{fig:tsne_results1}. We observe that PCA is able to effectively distinguish about 6 faults, while ConvAE increases this number up to 11. All the other faults are referred to a large, mixed group. In contrast, SensorSCAN not only distinguishes a larger number of faults, but it also constructs groups that almost entirely consist of only two faults. This potentially means that such a group can be effectively split into two by fine-tuning. Next, let us consider $\text{TEP}_{\text{Ricker}}$ in Figure \ref{fig:tsne_results2}. Again, we observe that SensorSCAN distinguishes a larger number of faults with respect to the competitors, while also constructing a mixed group with only two faults (the orange-blue group below the main mixed group).

\subsection{Semi-supervised setting}

We performed experiments in a setting that we call semi-supervised, in order to show the fine-tuning ability of SensorSCAN. We compared the results with the model based on Gated Recurrent Units (GRU) that was reported to show superior results on TEP in a supervised setting \cite{holy_grail}. 

First, we fine-tuned pretrained SensorSCAN and trained GRU on the dataset with a single labeled run to show that self-supervised pretraining benefits our model’s performance in the semi-supervised setting ($\text{SensorSCAN}_{\text{single}}$ and $\text{GRU}_\text{single}$, respectively – see Tables \ref{tab:benchmark1_semisupervised}, \ref{tab:benchmark2_semisupervised}). Second, we trained GRU on the full labeled dataset ($\text{GRU}_\text{full}$) and compared it with $\text{SensorSCAN}_{\text{single}}$ to illustrate that self-supervised pretraining and fine-tuning on a single run helps achieve performance close to a SOTA model trained on the full dataset.

\begin{table}[!ht]
    \centering
    \caption{FDD metrics evaluated on ${\text{TEP}_{\text{Rieth}}}$ in the semi-supervised setting. Top down: TPR/FPR for each fault, followed by aggregated detection and diagnosis metrics. The largest TPR values are highlighted on the condition that FPR is no greater than 0.05. The confidence interval is evaluated on randomly selected runs for training. SensorSCAN is fine-tuned on a single run. Results of GRU trained on a single run (left column) and on all runs (right column) are presented for comparison. Our fine-tuned model is either on a par with or is better than the single-run GRU and shows the best results on the number of faults that is similar to the all-runs GRU.}
    \label{tab:benchmark1_semisupervised}
    \begin{tabular}{p{25mm}llllll}
    \hline
         & \multicolumn{2}{l}{$\text{GRU}_\text{single}$} & \multicolumn{2}{l}{$\text{SensorSCAN}_\text{single}$} & \multicolumn{2}{l}{$\text{GRU}_\text{full}$} \\
         & TPR & FPR & TPR & FPR & TPR & FPR\\
    \hline
    Fault 1   & \textbf{1.00} $\pm$ 0.00 & 0.00 & \textbf{1.00} $\pm$ 0.00 & 0.00 & \textbf{1.00} & 0.00\\
    Fault 2   & \textbf{1.00} $\pm$ 0.00 & 0.00 & \textbf{1.00} $\pm$ 0.00 & 0.00 & \textbf{1.00} & 0.00\\
    Fault 3   & 0.15 $\pm$ 0.03 & 0.07 & \textbf{0.27} $\pm$ 0.04 & 0.05 & 0.79 & 0.13\\
    Fault 4   & 0.97 $\pm$ 0.01 & 0.00 & \textbf{1.00} $\pm$ 0.00 & 0.00 & \textbf{1.00} & 0.00\\
    Fault 5   & 0.93 $\pm$ 0.02 & 0.00 & \textbf{1.00} $\pm$ 0.00 & 0.00 & \textbf{1.00} & 0.00\\
    Fault 6   & 0.85 $\pm$ 0.03 & 0.00 & 0.97 $\pm$ 0.01 & 0.00 & \textbf{1.00} & 0.00\\
    Fault 7   & 0.99 $\pm$ 0.01 & 0.00 & \textbf{1.00} $\pm$ 0.00 & 0.00 & \textbf{1.00} & 0.00\\
    Fault 8   & 0.60 $\pm$ 0.05 & 0.00 & \textbf{0.90} $\pm$ 0.02 & 0.00 & 0.89 & 0.00\\
    Fault 9   & \textbf{0.03} $\pm$ 0.01 & 0.02 & 0.10 $\pm$ 0.04 & 0.08 & 0.00 & 0.00\\
    Fault 10   & 0.47 $\pm$ 0.19 & 0.01 & \textbf{0.68} $\pm$ 0.15 & 0.00 & 0.56 & 0.00\\
    Fault 11   & 0.99 $\pm$ 0.00 & 0.00 & \textbf{1.00} $\pm$ 0.00 & 0.00 & \textbf{1.00} & 0.00\\
    Fault 12   & 0.83 $\pm$ 0.03 & 0.00 & \textbf{0.96} $\pm$ 0.01 & 0.00 & 0.88 & 0.00\\
    Fault 13   & 0.30 $\pm$ 0.09 & 0.00 & \textbf{0.80} $\pm$ 0.03 & 0.00 & 0.70 & 0.00\\
    Fault 14   & \textbf{1.00} $\pm$ 0.00 & 0.00 & \textbf{1.00} $\pm$0.00 & 0.01 & \textbf{1.00} & 0.00\\
    Fault 15   & 0.04 $\pm$ 0.01 & 0.03 & 0.06 $\pm$ 0.03 & 0.04 & 0.00 & 0.00\\
    Fault 16   & 0.32 $\pm$ 0.15 & 0.08 & \textbf{0.60} $\pm$ 0.17 & 0.01 & 0.25 & 0.00\\
    Fault 17   & \textbf{1.00} $\pm$ 0.00 & 0.00 & \textbf{1.00} $\pm$ 0.00 & 0.00 & \textbf{1.00} & 0.00\\
    Fault 18   & 0.83 $\pm$ 0.02 & 0.00 & 0.72 $\pm$ 0.05 & 0.00 & \textbf{0.98} & 0.00\\
    Fault 19   & 0.99 $\pm$ 0.00 & 0.00 & \textbf{1.00} $\pm$ 0.00 & 0.00 &  \textbf{1.00} & 0.00\\
    Fault 20   & \textbf{1.00} $\pm$ 0.00 & 0.00 & \textbf{1.00} $\pm$ 0.00 & 0.00 &  \textbf{1.00} & 0.00\\
    
    \hline
    Detection TPR & \multicolumn{2}{c}{$\phantom{0}$0.85 $\pm$ 0.01} & \multicolumn{2}{c}{$\phantom{0}$\textbf{0.89} $\pm$ 0.01} & \multicolumn{2}{c}{0.87} \\
    Detection FPR & \multicolumn{2}{c}{$\phantom{0}$0.21 $\pm$ 0.06} & \multicolumn{2}{c}{$\phantom{0}$0.19 $\pm$ 0.02} & \multicolumn{2}{c}{\textbf{0.16}}\\
    CDR & \multicolumn{2}{c}{$\phantom{0}$0.84 $\pm$ 0.01} & \multicolumn{2}{c}{$\phantom{0}$0.89 $\pm$ 0.02} & \multicolumn{2}{c}{\textbf{0.92}}\\
    ADD & \multicolumn{2}{c}{30.78 $\pm$ 11.58} & \multicolumn{2}{c}{17.46 $\pm$ 3.08} & \multicolumn{2}{c}{\textbf{16.70}}\\
    \end{tabular}
\end{table}

\begin{table}[!ht]
    \centering
    \caption{FDD metrics evaluated on ${\text{TEP}_{\text{Ricker}}}$ in the semi-supervised setting. Top down: TPR/FPR for each fault, followed by aggregated detection and diagnosis metrics. The largest TPR values are highlighted on the condition that FPR is no greater than 0.05. The confidence interval is evaluated on randomly selected runs for training. SensorSCAN is fine-tuned on a single run. Results of GRU trained on a single run (left column) and on all runs (right column) are presented for comparison. Our fine-tuned model is either on par with or is better than the single-run GRU and shows the best results on the number of faults that is similar to the all-runs GRU.}
    \label{tab:benchmark2_semisupervised}
    \begin{tabular}{p{25mm}llllll}
    \hline
         & \multicolumn{2}{l}{$\text{GRU}_\text{single}$} & \multicolumn{2}{l}{$\text{SensorSCAN}_\text{single}$} & \multicolumn{2}{l}{$\text{GRU}_\text{full}$} \\
         & TPR & FPR & TPR & FPR & TPR & FPR\\
    \hline
    Fault 1    & 0.99 $\pm$ 0.00 & 0.00& \textbf{1.00} $\pm$ 0.00 & 0.00 & \textbf{1.00} & 0.00\\
    Fault 2    & 0.98 $\pm$ 0.00 & 0.00& \textbf{1.00} $\pm$ 0.00 & 0.00 & 0.99 & 0.00\\
    Fault 3    & 0.82 $\pm$ 0.12  & 0.00 & 0.97 $\pm$ 0.00 & 0.00 & \textbf{0.99} & 0.00\\
    Fault 4    & \textbf{1.00} $\pm$ 0.00 & 0.00 & \textbf{1.00} $\pm$ 0.00 & 0.00 & \textbf{1.00} & 0.00\\
    Fault 5    & 0.94 $\pm$ 0.01 & 0.01 & 0.99 $\pm$ 0.00 & 0.01 & \textbf{1.00} & 0.00\\
    Fault 6    & 0.99 $\pm$ 0.00 & 0.00 & \textbf{1.00} $\pm$ 0.00 & 0.00 & \textbf{1.00} & 0.00\\
    Fault 7    & \textbf{1.00} $\pm$ 0.00 & 0.00 & \textbf{1.00} $\pm$ 0.00 & 0.00 & \textbf{1.00} & 0.00\\
    Fault 8    & 0.87 $\pm$ 0.01 & 0.00 & 0.97 $\pm$ 0.01 & 0.00 & \textbf{0.99} & 0.00\\
    Fault 9    & 0.25 $\pm$ 0.20 & 0.01 & 0.54 $\pm$ 0.10 & 0.00 & \textbf{0.73} & 0.00\\
    Fault 10   & 0.95 $\pm$ 0.01 & 0.01 & \textbf{0.98} $\pm$ 0.00 & 0.00 & \textbf{0.98} & 0.00\\
    Fault 11   & 0.98 $\pm$ 0.01 & 0.00 & \textbf{0.99} $\pm$ 0.00 & 0.00 & \textbf{0.99} & 0.00\\
    Fault 12   & 0.95 $\pm$ 0.01 & 0.01 & \textbf{0.99} $\pm$ 0.00 & 0.00 & \textbf{0.99} & 0.00\\
    Fault 13   & 0.85 $\pm$ 0.01 & 0.00 & 0.97 $\pm$ 0.00 & 0.00 & \textbf{0.97} & 0.00\\
    Fault 14   & 0.99 $\pm$ 0.00 & 0.00 & \textbf{1.00} $\pm$ 0.00 & 0.00 & \textbf{1.00} & 0.00\\
    Fault 15   & 0.00 $\pm$ 0.01 & 0.02 & 0.06 $\pm$ 0.02 & 0.03 & \textbf{0.65} & 0.05\\
    Fault 16   & 0.02 $\pm$ 0.02 & 0.01 & 0.82 $\pm$ 0.04 & 0.01 & \textbf{0.95} & 0.01\\
    Fault 17   & 0.97 $\pm$ 0.00 & 0.00 & \textbf{0.98} $\pm$ 0.00 & 0.00 & \textbf{0.98} & 0.00\\
    Fault 18   & 0.96 $\pm$ 0.00 & 0.00 & 0.96 $\pm$ 0.00 & 0.00 & \textbf{0.97} & 0.00\\
    Fault 19   & 0.60 $\pm$ 0.16 & 0.00 & 0.89 $\pm$ 0.08 & 0.00 & \textbf{0.99} & 0.00\\
    Fault 20   & \textbf{0.97} $\pm$ 0.00 & 0.00 & \textbf{0.97} $\pm$ 0.00 & 0.01 & \textbf{0.97} & 0.00\\
    Fault 21   & 0.01 $\pm$ 0.02 & 0.00 & \textbf{0.06} $\pm$ 0.01 & 0.04 & 0.00 & 0.00\\
    Fault 22   & 0.25 $\pm$ 0.11 & 0.02 & 0.52 $\pm$ 0.13 & 0.01 & \textbf{0.74} & 0.00\\
    Fault 23   & 0.03 $\pm$ 0.04 & 0.00 & \textbf{0.98} $\pm$ 0.00 & 0.00 & \textbf{0.98} & 0.01\\
    Fault 24   & 0.96 $\pm$ 0.00 & 0.00 & \textbf{0.99} $\pm$ 0.00 & 0.00 & \textbf{0.99} & 0.00\\
    Fault 25   & 0.95 $\pm$ 0.00 & 0.00 & \textbf{0.99} $\pm$ 0.00 & 0.00 & \textbf{0.99} & 0.00\\
    Fault 26   & 0.89 $\pm$ 0.06 & 0.01 & \textbf{0.98} $\pm$ 0.00 & 0.00 & \textbf{0.98} & 0.00\\
    Fault 27   & 0.97 $\pm$ 0.01 & 0.00 & \textbf{0.99} $\pm$ 0.00 & 0.00 & \textbf{0.99} & 0.00\\
    Fault 28   & 0.28 $\pm$ 0.13 & 0.06 & 0.96 $\pm$ 0.00 & 0.01 & \textbf{0.97} & 0.01\\
    \hline
    Detection TPR & \multicolumn{2}{l}{$\phantom{0}$0.83 $\pm$ 0.02} & \multicolumn{2}{l}{$\phantom{0}$0.92 $\pm$ 0.00} & \multicolumn{2}{l}{\textbf{$\phantom{0}$0.94}}\\
    Detection FPR & \multicolumn{2}{l}{$\phantom{0}$0.11 $\pm$ 0.05} & \multicolumn{2}{l}{$\phantom{0}$0.09 $\pm$ 0.02} & \multicolumn{2}{l}{\textbf{$\phantom{0}$0.05}}\\
    CDR & \multicolumn{2}{l}{$\phantom{0}$0.88 $\pm$ 0.01} & \multicolumn{2}{l}{$\phantom{0}$0.95 $\pm$ 0.00} & \multicolumn{2}{l}{\textbf{$\phantom{0}$0.97}}\\
    ADD & \multicolumn{2}{l}{59.76 $\pm$ 18.49} & \multicolumn{2}{l}{35.47  $\pm$ 1.59} & \multicolumn{2}{l}{\textbf{31.00}}\\
    \hline
    \end{tabular}
\end{table}

Remarkably, GRU with a limited number of labeled samples is disadvantaged over our approach since our model has access to all sensor data (for the semi-supervised pretraining part), whereas GRU only uses the labeled sensor data. We can observe that SensorSCAN is better than the single-run GRU in detecting most faults and is at least as good in the other faults. In addition, our model’s values are close to those of the all-runs GRU, which is valid for both datasets. SensorSCAN showed the highest TPR in 14 faults, while the all-runs GRU proved superior in 15 faults in $\text{TEP}_\text{Rieth}$ (and 17 vs. 26 in $\text{TEP}_\text{Ricker}$). Important observations are to be made on the ability of the models to handle the hard-to-detect faults. Notably, only the all-runs GRU successfully detected fault 9 in $\text{TEP}_{\text{Rieth}}$, while fault 15 persisted across all the models: the single-run GRU did not detect it, whereas the others yielded inadequately high FPR (0.10 for SensorSCAN, 0.11 for the all-runs GRU) -- assuming that FPR greater than 0.05 is impractical in real cases due to unacceptably frequent false alarms. In $\text{TEP}_{\text{Ricker}}$, fault 9 was now detected by all the models, with the all-runs GRU showing the highest TPR. Fault 15 was successfully detected by the all-runs GRU, while the others showed inadequately low TPR (0.01 for the single-run GRU and 0.04 for SensorSCAN). Fault 16 was detected by SensorSCAN and the all-runs GRU, while the single-run GRU produced the highest TPR. Fault 21 remained almost undetected by all the models: only SensorSCAN was able to detect a few samples (with the TPR of 0.04). The hard-to-detect faults significantly increase Detection FPR, thus making almost all of the models trained in the semi-supervised setting impractical in real-world cases. Only the all-runs GRU resulted with the threshold Detection FPR value of 0.05 on $\text{TEP}_{\text{Ricker}}$. Nevertheless, we can conclude that fine-tuning helps SensorSCAN not only achieve results similar to the all-runs GRU but also to cope with the hard-to-detect faults 9 and 16 in $\text{TEP}_{\text{Ricker}}$. In real cases, we can decrease Detection FPR of SensorSCAN to acceptable values by turning off the detection of the remaining hard-to-detect faults and treating them with other methods -- for example, the methods based on expert knowledge.

\section{Discussion}
\label{discussion}

The experimental results presented in Figure \ref{fig:polar_fdd} demonstrate the ability of our model to outperform the latest deep learning data-driven FDD models in the unsupervised setting. Note that our model is able to successfully detect the faults that remain undetected by other models. It becomes possible thanks to the powerful feature extractor that can retrieve inherent patterns from the process behavior and map them as separate groups in the latent space. Subsequent deep clustering increases the density of such groups, which forces the feature extractor not only to map the process patterns in the latent space but also to distinguish between them, making the diagnostics more accurate. 

As can be seen in Table \ref{tab:benchmark1_unsupervised_agg} and Table \ref{tab:benchmark2_unsupervised_agg}, SensorSCAN detects faults much faster in the unsupervised setting in terms of ADD. A large ADD may occur in the other models because of smooth transitions between process states in the latent space. Samples that are uniformly distant from the cluster centers increase the uncertainty of the model and do not allow timely fault detection. We showed that the deep clustering approach used in our model makes groups of samples denser, thus reducing the uncertainty of the model, which subsequently leads to an abrupt transition between the predicted process states, followed by undelayed detection. 

Considering the application of FDD models in real-world industrial cases, it is important to assess the computational time and the memory costs. Deep-learning methods are more expensive compared to classical methods like PCA, but this is mainly due to the training process, since it requires performing a complicated backpropagation procedure and implementing optimization steps over a large number of samples with stochastic gradient descent (SGD) or similar methods.

The time complexity of the training process of deep learning models depends on many factors, such as the size of the dataset, the size of mini-batches, the learning rate, the number of steps, and the optimization algorithm. Let us look at the difficulty of training such deep neural networks if the number of sensors significantly increases, which is a realistic scenario for large-scale complex processes. The number of sensors determines only the number of neurons in the input layer in a deep neural network. That is, increasing the number of sensors does not substantially affect the time complexity of a single optimization step. For example, increasing the number of sensors from 52 to 1000 in SensorSCAN leads to the increase in the execution time of one optimization step with the mini-batch size of 512 from $0.46 \pm 0.03$ seconds to $0.49 \pm 0.04$ seconds.

Time costs in the inference are low since the samples are received in the one-by-one fashion with an intermediate delay. For example, prediction of the process state of a single sample by SensorSCAN takes approximately 0.01 seconds on a laptop with the 2.7 GHz Dual-Core Intel Core i5 CPU, while samples arrive every 3 minutes in TEP. As a result, we can  process each sample without any delay. 

Memory costs depend on the number of trainable parameters; the exact values (in megabytes) used in our experiments are represented in Table \ref{tab:memory_costs}.

\begin{table}[!ht]
    \centering
    \caption{Memory costs of models on TEP dataset.}
    \label{tab:memory_costs}
    \begin{tabular}{lc}
    \hline
         &  MB\\
    \hline
    PCA             & 0.41\\
    ST-CatGAN       & 0.82\\
    ConvAE          & 4.97\\
    Ours            & 2.38\\
    \hline
    \end{tabular}
\end{table}

Summing up, we conclude that usage of deep-learning FDD models is possible in real-time monitoring even on a laptop with CPU cores.

\section{Conclusion}
\label{conclusion}

In this paper, we proposed an unsupervised FDD model based on self-supervised learning and deep clustering. The predicted faults were obtained from cluster indices by means of label matching procedure, a simple heuristic technique that in real-world industrial setting can be performed by experts. The model was evaluated on the datasets simulated by Tennessee Eastman Process benchmarks, $\text{TEP}_\text{Rieth}$ \cite{extended_tep} and $\text{TEP}_\text{Ricker}$ \cite{new_ext_tep}. We evaluated our model using a wide range of metrics assessing  the ability to cluster faults in the latent space, the ability to detect different faults, and the speed of detection. The empirical evaluation  showed that our method outperforms other unsupervised approaches on both datasets due to the ability of the feature extractor to represent samples of sensor data in the latent space while preserving important high-level properties and the ability of deep clustering to produce dense clusters in the latent space.

Data with insufficient labeling of faults or even without labeling is common in the real-world industrial setting. Most of the existing FDD approaches are supervised \cite{survey1, survey2}, which basically ignores this fact and makes their application impractical. SensorSCAN is based on SSL and deep clustering, which enables training on unlabeled data and taking into account the expected number of faults, while fine-tuning allows us to take advantage of the  labeled examples in the data. Moreover, if the expected number of faults is changed or new faults are incorporated, then the feature extractor can be further fine-tuned without needing to train it from scratch. Thus, our method not only advances the state-of-the-art metrics on the chosen datasets but also covers the machine learning settings used in real-world monitoring and control of industrial processes, which is essential for adapting data-driven models to the production stage.

We showed that our model can be applied in the case of unlabeled data, which is a typical scenario in real industrial processes. However, we observed that some faults consistently resist  detection in the absence of labels; to address this obstacle, we proposed semi-supervised fine-tuning, the technique that helps detect difficult faults using very few labeled data. 

We envisage further research in two directions. The first is to explore other SSL techniques, including the methods that incorporate knowledge from the domains of chemistry and physics. The second is the analysis of deep semi-supervised methods for performing fine-tuning. We think that the promising results recently achieved by semi-supervised methods for time series classification \cite{semitime, tapnet, semisupervised_ts} can be adapted for FDD on sensor data.

\section*{Acknowledgements}

The work on the Related work and the Ablation sections was supported by the Russian Science Foundation under grant 22-11-00323 and performed at the National Research University Higher School of Economics (Moscow, Russia).

\bibliographystyle{elsarticle-num}
\bibliography{refs}

\newpage
\appendix

\section{Detailed description of Tennessee Eastman Process}
\label{tep_details}

The Tennessee Eastman Process consists of five major units (reactor, product condenser, vapor-liquid separator, recycle compressor, and product stripper) to produce liquids G and H from gaseous reactants A, C, D, E with inert B and byproduct F. The process's reactions are exothermic and irreversible and are described by:

\begin{center}
    \begin{tabular}{rccc}
    A(gas) + C(gas) + D(gas) & $\longrightarrow$ & G(liq), & Product 1, \\
    A(gas) + C(gas) + E(gas) & $\longrightarrow$ & H(liq), & Product 2, \\
    A(gas)  + E(gas) & $\longrightarrow$ & F(liq), & Byproduct, \\
    3D(gas) & $\longrightarrow$ & 2F(liq), & Byproduct. \\
    \end{tabular}
\end{center}
    
There are 52 process variables: 22 continuously measured  (Table \ref{tab:cont}), 19 sampled measured (Table \ref{tab:sampled}) and 11 manipulated  variables (Table \ref{tab:manipulated}). The measured variables are affected by measurement noise and the sampled variables are obtained with a certain delay. The process undergoes 20 predefined faults (Table \ref{tab:faults}) with 15 knowns and 5 unknowns.

Faults 1-7 are associated with the step change in a process variable, faults 8-12 are related to increased variability of specific process variables, fault 13 is a slow drift in reaction kinetics, and faults 14 and 15 are caused by sticking valves.

To generate historical data, the process is simulated with a control scheme. Originally, the TEP simulation presented by Eastman Chemical Company was available in open-loop operation. However, the TEP process is open-loop unstable even with initialization from \cite{optimal_steady}. Besides, real industrial plants operate in a closed loop, so the employment of a closed-loop control system seems natural. There are six process operation modes (see Table \ref{tab:modes}) which correspond to various G/H mass ratios and production rates in Stream 11.

\begin{table}[H]
    \centering
    \caption{Process operation modes proposed in \cite{original_tep}.}
    \label{tab:modes}
    \begin{tabular}{ccc}
    \hline
    Mode     & G/H mass ratio & G production rate (stream 11) \\
    \hline
    1     &  50/50  & 7038 kg h\textsuperscript{-1}\\
    2     &  10/90  & 1408 kg h\textsuperscript{-1}\\
    3     &  90/10  & 10000 kg h\textsuperscript{-1}\\
    4     &  50/50  &  maximum production rate \\
    5     &  10/90  &  maximum production rate  \\
    6     &  90/10  &  maximum production rate  \\
    \hline
    \end{tabular}
\end{table}

\begin{figure}[H]
    \centering
    \includegraphics[width=1\linewidth]{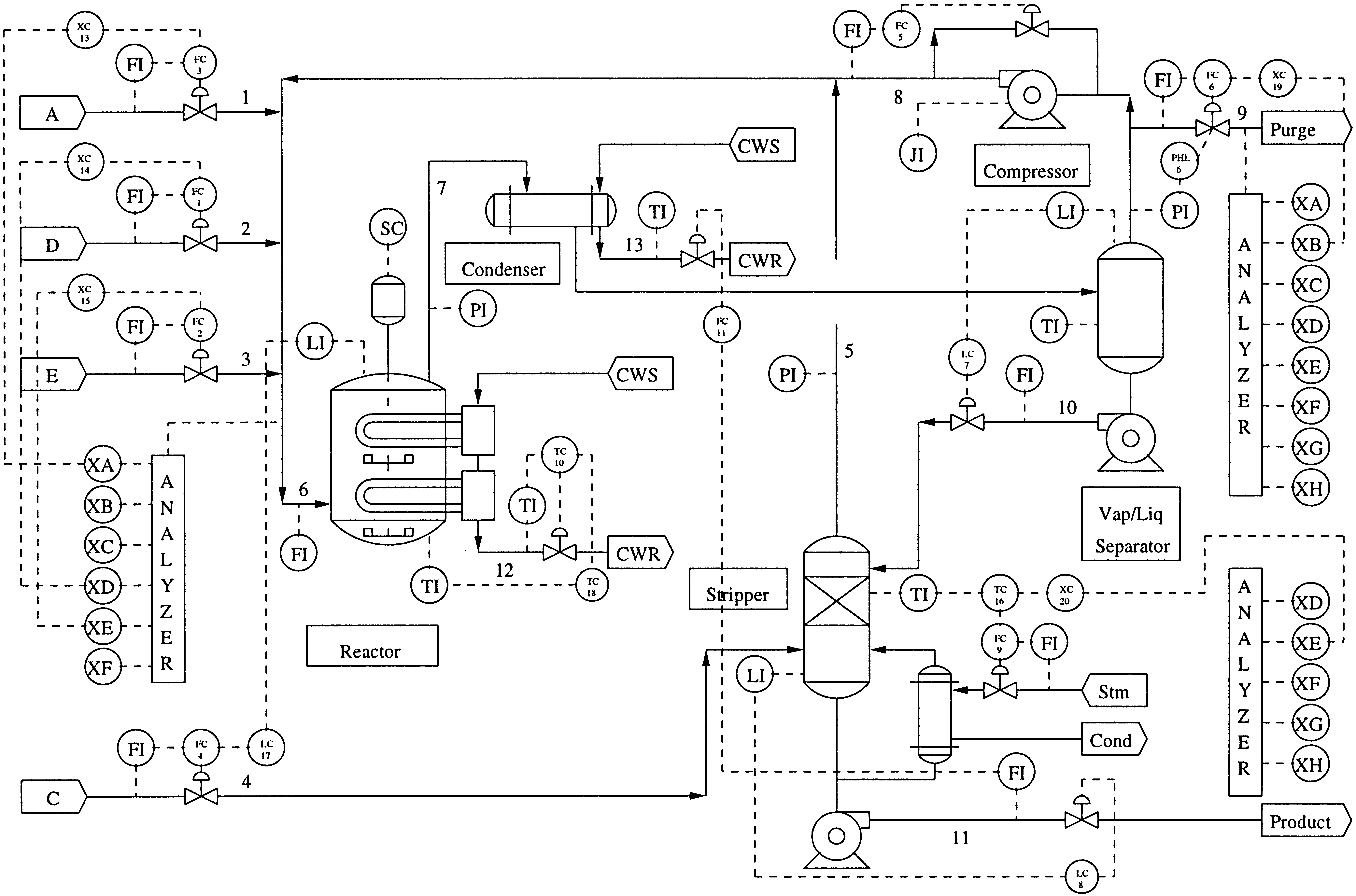}
    \caption{Diagram of the Tennessee Eastman Process simulator \cite{data_simulator}.}
    \label{fig:diagram}
\end{figure}

\begin{table}[H]
    \centering
    \caption{Continuous process measurements.}
    \label{tab:cont}
    \begin{tabular}{ccc}
    \hline
    Variable     &  Description & Units \\
    \hline
    XMEAS(1)     & A feed (stream 1)  & ks cm h \\
    XMEAS(2)     & D feed (stream 2)  & kg h\textsuperscript{-1} \\
    XMEAS(3)     & E feed (stream 3)  & kg h\textsuperscript{-1} \\
    XMEAS(4)     & A and C feed (stream 4)  & ks cm h \\
    XMEAS(5)     & Recycle flow (stream 8)  & ks cm h \\
    XMEAS(6)     & Reactor feed rate (stream 6)  & ks cm h \\
    XMEAS(7)     & Reactor pressure  & kPa gauge \\
    XMEAS(8)     & Reactor level  & \% \\
    XMEAS(9)     & Reactor temperature  & \degree C \\
    XMEAS(10)     & Purge rate (stream 9)  & ks cm h \\
    XMEAS(11)    & Product separator temperature & \degree C \\
    XMEAS(12)    & Product separator level &  \% \\
    XMEAS(13)    & Product separator pressure &   kPa gauge \\
    XMEAS(14)    & Product separator underflow (stream 10) &  m\textsuperscript{3 } h\textsuperscript{-1}\\
    XMEAS(15)    & Stripper level & \%  \\
    XMEAS(16)    & Stripper pressure &  kPa gauge  \\
    XMEAS(17)    & Stripper underflow (stream 11) &  m\textsuperscript{3} h\textsuperscript{-1}  \\
    XMEAS(18)    & Stripper temperature & \degree C  \\
    XMEAS(19)    & Stripper steam flow &  kg h\textsuperscript{-1} \\
    XMEAS(20)    & Compressor work & \degree C  \\
    XMEAS(21)    & Reactor cooling water outlet temperature & \degree C  \\
    XMEAS(22)    & Stripper temperature & \degree C  \\
    \hline
    \end{tabular}
\end{table}

\begin{table}[H]
    \centering
    \caption{Sampled process measurements, in mole \%}
    \label{tab:sampled}
    \begin{tabular}{ccc}
    \hline
    Block     & Variable & Description\\
    \hline
     Reactor feed analysis    & XMEAS(23) &  Component A\\
     & XMEAS(24) & Component B\\
      & XMEAS(25) & Component C\\
       & XMEAS(26) & Component D\\
        & XMEAS(27) & Component E\\
         & XMEAS(28) & Component F\\
         
    Purge gas analysis    & XMEAS(29) &  Component A\\
    & XMEAS(30) & Component B\\
      & XMEAS(31) & Component C\\
       & XMEAS(32) & Component D\\
        & XMEAS(33) & Component E\\
         & XMEAS(34) & Component F\\
         
    & XMEAS(35) & Component G\\
    & XMEAS(36) & Component H\\
    
    Product analysis    & XMEAS(37) &  Component D\\
    & XMEAS(38) & Component E\\
    & XMEAS(39) & Component F\\
    & XMEAS(40) & Component G\\
    & XMEAS(41) & Component H\\
    \hline
    \end{tabular}
\end{table}

\begin{table}[H]
    \centering
    \caption{Manipulated variables.}
    \label{tab:manipulated}
    \begin{tabular}{ccc}
    \hline
    Variable     &  Description & Units \\
    \hline
    XMV(1)     & D feed flow (stream 2) & kg h\textsuperscript{-1} \\
    XMV(2)     & E feed flow (stream 3) & kg h\textsuperscript{-1} \\
    XMV(3)     & A feed flow (stream 1) & ks cm h \\
    XMV(4)     & A and C feed flow (stream 4) & ks cm h \\
    XMV(5)     & Compressor recycle valve & \% \\
    XMV(6)     & D feed flow (stream 2) & \% \\
    XMV(7)     & D feed flow (stream 2) & m\textsuperscript{3} h\textsuperscript{-1} \\
    XMV(8)     & D feed flow (stream 2) &  m\textsuperscript{3} h\textsuperscript{-1} \\
    XMV(9)     & D feed flow (stream 2) & \% \\
    XMV(10)     & D feed flow (stream 2) &  m\textsuperscript{3} h\textsuperscript{-1} \\
    XMV(11)     & D feed flow (stream 2) &  m\textsuperscript{3} h\textsuperscript{-1} \\
    \hline
    \end{tabular}
\end{table}

\begin{table}[H]
    \centering
    \caption{Process faults. Faults introduced in \cite{revision} are located under the dashed line.}
    \label{tab:faults}
    \begin{tabular}{ccc}
    \hline
    Fault number     & Process variable & Type \\
    \hline
    IDV(1)     & A/C feed ratio, B composition constant (stream 4) & Step\\
    IDV(2)     &  B composition, A/C ration constant (stream 4) & Step\\
    IDV(3)     &  D feed temperature (stream 2) & Step\\
    IDV(4)     &  Reactor cooling water inlet temperature & Step\\
    IDV(5)     & Condenser cooling water inlet temperature  & Step\\
    IDV(6)     & A feed loss (stream 1) & Step\\
    IDV(7)     & C header pressure loss - reduced availability & Step \\
    IDV(8)     & A, B, C feed composition (stream 4) & Random variation\\
    IDV(9)     & D feed temperature (stream 2) & Random variation\\
    IDV(10)     & C feed temperature (stream 4) & Random variation\\
    IDV(11)     & Reactor cooling water inlet temperature & Random variation\\
    IDV(12)     & Condenser cooling water inlet temperature  & Random variation\\
    IDV(13)     & Reaction kinetics & Slow drift \\
    IDV(14)     & Reactor cooling water valve & Sticking \\
    IDV(15)     & Condencer cooling water valve & Sticking \\
    IDV(16)     & Unknown  & Unknown\\
    IDV(17)     & Unknown & Unknown\\
    IDV(18)     & Unknown & Unknown\\
    IDV(19)     & Unknown  & Unknown\\
    IDV(20)     & Unknown & Random variation\\
    \hdashline
    IDV(21)     & A feed (stream 1) temperature & Random variation\\
    IDV(22)     & E feed (stream 3) temperature & Random variation\\
    IDV(23)     & A feed flow (stream 1) & Random variation\\
    IDV(24)     & D feed flow (stream 2) & Random variation\\
    IDV(25)     & E feed flow (stream 3) & Random variation\\
    IDV(26)     & A and C feed flow (stream 4) & Random variation\\
    IDV(27)     & Reactor cooling water flow & Random variation\\
    IDV(28)     & Condenser cooling water flow & Random variation\\
    \hline
    \end{tabular}
\end{table}

\newpage
\section{Training setup for SensorSCAN}
\label{appendix_training}

The neural network is implemented with the PyTorch \cite{pytorch} library for Python. We initialize the weights with Xavier initialization and perform optimization with Adam \cite{adam} for all models. We use the NVIDIA Tesla A100 80Gb for calculations. The hyperparameters are selected using grid search.

To generate the training dataset, we use the sliding window size of L equal to 100 and the step size of 1 for both datasets. The sensor data is normalized with feature-wise standardization (for every feature, the mean and the scale are calculated over all timesteps). We follow \cite{unsupervised_tep} in utilizing only a subset of process measurements. We removed 19 sampled process measurements and the measurements that remain constant. As a result, the overall number of sensors D totals 30.

\subsection{First step}

We employ only the training dataset for feature extractor pretraining. The best combination of hyperparameters is picked via visual evaluation of the training dataset projection onto the embedding space produced with t-SNE without fault-based coloring since the latter is not available in the real world scenario. The smoother and more non-intersecting clusters are indicative of the  optimal combination of hyperparameters.

The training takes $E = 8$ epochs. The learning rate is equal to $10^{-3}$ , the batch size B is equal to 1024 and the weight decay is equal to $10^{-4}$. We found that the large batch size is critical for achieving peak performance. Surprisingly, this finding is connected with the reconstruction task but not with contrastive learning. Usually, contrastive learning requires utilization of a large minibatch.

The other hyperparameters are set as follows: $\lambda_{\text{cont}} = 0.7$ (see Eq. \ref{sum_loss}), $\tau = 0.2$ (see Eq. \ref{contrastive_loss}), $r$ = 0.5, and $l_m = 6$ (see Eq. \ref{mask_length}). We conducted experiments with gradually increasing the length of the mask $l_m$ and the masking ratio $r$ but have not noticed a statistically significant difference. Regarding the architecture of the model, the encoder embedding dimension H is set to 128, the feed-forward layer dimension is 512, the projection head dimension F is 32, and the dropout rate is equal to 0.1.

For the augmentations, the following hyperparameters are found with grid search. For jitter augmentation, the noise is sampled from a normal distribution with the zero mean and the standard deviation equaling 0.08. For scaling augmentation, the scaling factor is also sampled from the normal distribution with the standard deviation set to 0.1 and the mean equal to 2 and 0.5 for weak and strong augmentations, respectively. During hyperparameter tuning, we found that the effectiveness of learning significantly depends on the number of chunks into which the time series is divided for permutation. Splitting into about 15 chunks exhibits superior results, which is quite counterintuitive since such transformation seriously distorts the temporal information. Presumably, this happens because a large number of chunks induces greater variability in augmentations, which makes the contrastive learning task difficult enough for the model to learn meaningful features.

\subsection{Second step}

The model is trained for 5 epochs. The weights of the feature extractor are frozen for three epochs to avoid distortions since the clustering network is randomly initialized and the learning process is quite unstable at the beginning. The best combination of hyperparameters is chosen according to the value of SCAN loss, which does not employ ground truth labels. This way, both steps do not use information about the ground truth fault distribution. Therefore, we consider our hyperparameter tuning approach fair. The learning rate is equal to 1e-2 for the clustering network and to 4e-5 for the feature extractor. The batch size is set to be 128, $\lambda_{ent}$ is equal to 2, the number of neighbors K is equal to 12, and the number of chunks T is 20.

\newpage
\section{FDD metrics}
\label{appendix_metrics}

This section provides the TPR and FPR values from the radar charts in Sections \ref{ablation_study} and \ref{results}. Tables \ref{tab:benchmark2_clustersize} and \ref{tab:benchmark2_faults_ablation} show the results of the ablation study for selection of the number of clusters and of the fault subset, respectively. Tables \ref{tab:benchmark1_unsupervised} and \ref{tab:benchmark2_unsupervised} show the results of the methods evaluation in the unsupervised setting on ${\text{TEP}_{\text{Rieth}}}$ and ${\text{TEP}_{\text{Ricker}}}$, respectively.

\begin{table}[!ht]
    \centering
    \caption{FDD metrics evaluated on ${\text{TEP}_{\text{Ricker}}}$ for various numbers of clusters. Top down: TPR/FPR for each fault, followed by aggregated detection and diagnosis metrics. The largest TPR values are highlighted on the condition that FPR is not greater than 0.05.}
    \label{tab:benchmark2_clustersize}
    \begin{tabular}{lcccc}
    \hline
         &  10 & 29 & 33 & 58\\
    \hline
    Fault 1 &  0.00/0.00 & \textbf{0.98}/0.00 & 0.87/0.00 & 0.65/0.00\\
    Fault 2 &  0.99/0.00 & 0.99/0.00 & 0.87/0.00 & \textbf{1.00}/0.00\\
    Fault 3 &  0.00/0.00 & \textbf{0.95}/0.00 & 0.77/0.00 & \textbf{0.95}/0.00\\
    Fault 4 &  0.99/0.00 & \textbf{1.00}/0.00 & 0.93/0.00 & 0.59/0.00\\
    Fault 5 &  0.99/0.00 & \textbf{1.00}/0.00 & 0.98/0.00 & 0.99/0.00\\
    Fault 6 &  0.98/0.00 & \textbf{0.99}/0.00 & 0.93/0.00 & \textbf{0.99}/0.00\\
    Fault 7 &  0.00/0.00 & \textbf{0.93}/0.00 & \textbf{0.93}/0.00 & \textbf{0.93}/0.00\\
    Fault 8 &  0.00/0.00 & \textbf{0.99}/0.00 & 0.92/0.00 & 0.92/0.00\\
    Fault 9 &  0.00/0.00 & 0.00/0.00 & \textbf{0.53}/0.00 & 0.27/0.00\\
    Fault 10 & 0.00/0.00 & \textbf{0.98}/0.00 & 0.92/0.00 & 0.78/0.00\\
    Fault 11 & 0.93/0.00 & \textbf{0.99}/0.00 & 0.93/0.00 & 0.93/0.00\\
    Fault 12 & 0.00/0.00 & \textbf{0.99}/0.00 & 0.93/0.00 & 0.96/0.00\\
    Fault 13 & 0.00/0.00 & \textbf{0.96}/0.00 & 0.91/0.00 & 0.88/0.00\\
    Fault 14 & \textbf{1.00}/0.00 & \textbf{1.00}/0.00 & 0.93/0.00 & 0.93/0.00\\
    Fault 15 & 0.00/0.00 & 0.00/0.00 & 0.00/0.00 & 0.00/0.00\\
    Fault 16 & 0.00/0.00 & 0.00/0.00 & \textbf{0.75}/0.00 & 0.69/0.01\\
    Fault 17 & 0.00/0.00 & \textbf{0.91}/0.00 & \textbf{0.91}/0.00 & \textbf{0.91}/0.00\\
    Fault 18 & 0.00/0.00 & \textbf{0.96}/0.00 & \textbf{0.96}/0.00 & \textbf{0.96}/0.00\\
    Fault 19 & 0.00/0.00 & \textbf{0.99}/0.00 & 0.98/0.00 & 0.89/0.00\\
    Fault 20 & 0.00/0.00 & \textbf{0.97}/0.00 & \textbf{0.97}/0.00 & \textbf{0.97}/0.00\\
    Fault 21 & 0.00/0.00 & 0.00/0.00 & 0.00/0.00 & 0.00/0.00\\
    Fault 22 & \textbf{0.96}/0.00 & \textbf{0.96}/0.00 & 0.44/0.00 & 0.59/0.00\\
    Fault 23 & 0.00/0.00 & \textbf{0.96}/0.00 & 0.91/0.00 & 0.58/0.01\\
    Fault 24 & 0.00/0.00 & \textbf{0.99}/0.00 & 0.92/0.00 & 0.92/0.00\\
    Fault 25 & 0.00/0.00 & \textbf{0.95}/0.00 & 0.92/0.00 & 0.85/0.00\\
    Fault 26 & 0.97/0.00 & \textbf{0.98}/0.00 & 0.90/0.00 & 0.97/0.00\\
    Fault 27 & 0.00/0.00 & \textbf{0.99}/0.00 & 0.94/0.00 & 0.93/0.00\\
    Fault 28 & 0.00/0.00 & \textbf{0.96}/0.01 & 0.95/0.00 & \textbf{0.96}/0.01\\
    \hline
    \end{tabular}
\end{table}

\begin{table}[!ht]
    \centering
    \caption{FDD metrics evaluated on ${\text{TEP}_{\text{Ricker}}}$ for feature extractor pretraining with various faults. Top down: TPR/FPR for each fault. The largest TPR values are highlighted on the condition that FPR is no more than 0.05.}
    \label{tab:benchmark2_faults_ablation}
    \begin{tabular}{p{25mm}cccc}
    \hline
         & Untrained model & Easy faults & Difficult faults & All faults (ours) \\
    \hline
    Fault 1   & 0.90/0.00 & 0.84/0.00 & 0.89/0.00 & \textbf{0.98}/0.00 \\
    Fault 2   & 0.89/0.00 & 0.74/0.00 & 0.92/0.00 & \textbf{0.99}/0.00 \\
    Fault 3   & 0.00/0.00 & 0.00/0.00 & 0.00/0.00 & \textbf{0.95}/0.00 \\
    Fault 4   & 0.98/0.00 & 0.93/0.00 & 0.97/0.00 & \textbf{1.00}/0.00 \\
    Fault 5   & 0.96/0.00 & 0.96/0.00 & 0.00/0.00 & \textbf{1.00}/0.00 \\
    Fault 6   & 0.97/0.00 & 0.93/0.00 & 0.98/0.00 & \textbf{0.99}/0.00 \\
    Fault 7   & \textbf{1.00}/0.00 & 0.93/0.00 & 0.94/0.01 & 0.93/0.00 \\
    Fault 8   & 0.42/0.00 & 0.89/0.00 & 0.89/0.00 & \textbf{0.99}/0.00 \\
    Fault 9   & 0.00/0.00 & 0.00/0.00 & 0.00/0.00 & 0.00/0.00 \\
    Fault 10   & 0.60/0.00 & 0.94/0.00 & 0.97/0.01 & \textbf{0.98}/0.00 \\
    Fault 11   & 0.98/0.00 & 0.93/0.00 & 0.94/0.00 & \textbf{0.99}/0.00 \\
    Fault 12   & 0.00/0.00 & 0.91/0.00 & 0.92/0.00 & \textbf{0.99}/0.00 \\
    Fault 13   & 0.50/0.00 & 0.88/0.00 & 0.92/0.00 & \textbf{0.96}/0.00 \\
    Fault 14   & 0.98/0.00 & 0.52/0.00 & 0.96/0.00 & \textbf{1.00}/0.00 \\
    Fault 15   & 0.00/0.00 & 0.00/0.00 & 0.00/0.00 & 0.00/0.00 \\
    Fault 16   & 0.00/0.00 & 0.00/0.00 & 0.00/0.00 & 0.00/0.00 \\
    Fault 17   & \textbf{0.96}/0.00 & 0.94/0.00 & 0.92/0.00 & 0.91/0.00 \\
    Fault 18   & 0.94/0.00 & 0.93/0.00 & 0.95/0.00 & \textbf{0.96}/0.00 \\
    Fault 19   & 0.49/0.00 & 0.55/0.00 & 0.96/0.00 & \textbf{0.99}/0.00 \\
    Fault 20   & 0.96/0.00 & 0.95/0.00 & 0.95/0.00 & \textbf{0.97}/0.00 \\
    Fault 21   & 0.00/0.00 & 0.00/0.00 & 0.00/0.00 & 0.00/0.00 \\
    Fault 22   & 0.00/0.00 & 0.00/0.00 & 0.56/0.00 & \textbf{0.96}/0.00 \\
    Fault 23   & 0.00/0.00 & 0.00/0.00 & 0.00/0.00 & \textbf{0.96}/0.00 \\
    Fault 24   & 0.98/0.00 & 0.93/0.00 & 0.97/0.00 & \textbf{0.99}/0.00 \\
    Fault 25   & 0.96/0.00 & 0.90/0.00 & \textbf{0.98}/0.00 & 0.95/0.00 \\
    Fault 26   & 0.00/0.00 & 0.55/0.00 & 0.00/0.00 & \textbf{0.98}/0.00 \\
    Fault 27   & 0.00/0.00 & 0.55/0.00 & 0.93/0.00 & \textbf{0.99}/0.00 \\
    Fault 28   & 0.00/0.00 & 0.00/0.00 & 0.00/0.00 & \textbf{0.96}/0.01 \\
    \hline
    \end{tabular}
\end{table}

\begin{table}[!ht]
    \centering
    \caption{FDD metrics evaluated on ${\text{TEP}_{\text{Rieth}}}$ in the unsupervised setting. Top down: TPR/FPR for each fault, followed by aggregated detection and diagnosis metrics. The largest TPR values are highlighted on the condition that FPR is not greater than 0.05}
    \label{tab:benchmark1_unsupervised}
    \begin{tabular}{p{25mm}cccc}
    \hline
         &  PCA & ST-CatGAN  & ConvAE & Ours\\
    \hline
    Fault 1   & 0.92/0.00 & 0.00/0.00 & 0.95/0.00 & \textbf{1.00}/0.00\\
    Fault 2   & 0.88/0.00 & 0.85/0.00 & 0.95/0.00 & \textbf{1.00}/0.00\\
    Fault 3   & 0.00/0.00 & 0.00/0.00 & 0.00/0.00 & 0.00/0.00\\
    Fault 4   & 0.00/0.00 & 0.00/0.00 & 0.00/0.00 & \textbf{0.97}/0.00\\
    Fault 5   & 0.00/0.00 & 0.13/0.00 & 0.07/0.00 & \textbf{1.00}/0.00\\
    Fault 6   & 0.86/0.00 & 0.99/0.00 & 0.81/0.00 & \textbf{0.96}/0.00\\
    Fault 7   & 0.74/0.00 & 0.00/0.00 & 0.94/0.00 & \textbf{1.00}/0.00\\
    Fault 8   & 0.00/0.00 & 0.00/0.00 & 0.67/0.00 & \textbf{0.78}/0.00\\
    Fault 9   & 0.00/0.00 & 0.00/0.00 & 0.00/0.00 & 0.00/0.00\\
    Fault 10   & 0.00/0.00 & 0.00/0.00 & 0.00/0.00 & \textbf{0.69}/0.00\\
    Fault 11   & 0.00/0.00 & 0.00/0.00 & 0.00/0.00 & \textbf{1.00}/0.00\\
    Fault 12   & 0.37/0.00 & 0.00/0.00 & 0.08/0.00 & \textbf{0.95}/0.00\\
    Fault 13   & 0.00/0.00 & 0.00/0.00 & 0.38/0.00 & \textbf{0.76}/0.00\\
    Fault 14   & 0.33/0.00 & 0.00/0.00 & 0.00/0.00 & \textbf{0.98}/0.00\\
    Fault 15   & 0.00/0.00 & 0.00/0.00 & 0.00/0.00 & 0.00/0.00\\
    Fault 16   & 0.00/0.00 & 0.00/0.00 & 0.00/0.00 & \textbf{0.71}/0.00\\
    Fault 17   & 0.89/0.00 & 0.00/0.00 & 0.92/0.00 & \textbf{1.00}/0.00\\
    Fault 18   & 0.65/0.00 & 0.00/0.00 & 0.40/0.00 & \textbf{0.69}/0.00\\
    Fault 19   & 0.00/0.00 & 0.00/0.00 & 0.00/0.00 & \textbf{0.98}/0.00\\
    Fault 20   & 0.00/0.00 & 0.00/0.00 & 0.00/0.00 & \textbf{1.00}/0.00\\
    
    \hline
    \end{tabular}
\end{table}

\begin{table}[!ht]
    \centering
    \caption{FDD metrics evaluated on ${\text{TEP}_{\text{Ricker}}}$ in the unsupervised setting. Top down: TPR/FPR for each fault, then aggregated detection and diagnosis metrics. The largest TPR values are highlighted on the condition that FPR is not greater than 0.05.}
    \label{tab:benchmark2_unsupervised}
    \begin{tabular}{p{25mm}cccc}
    \hline
         &  PCA & ST-CatGAN  & ConvAE & Ours \\
    \hline
    Fault 1   & 0.90/0.00 & 0.88/0.00 & 0.79/0.00 & \textbf{0.98}/0.00 \\
    Fault 2   & 0.93/0.00 & 0.97/0.00 & 0.93/0.00 & \textbf{0.99}/0.00 \\
    Fault 3   & 0.00/0.00 & 0.00/0.00 & 0.00/0.00 & \textbf{0.95}/0.00 \\
    Fault 4   & 0.96/0.00 & 0.00/0.00 & 0.70/0.00 & \textbf{1.00}/0.00 \\
    Fault 5   & 0.00/0.00 & 0.00/0.00 & 0.97/0.00 & \textbf{1.00}/0.00 \\
    Fault 6   & 0.93/0.00 & 0.97/0.00 & 0.89/0.00 & \textbf{0.99}/0.00 \\
    Fault 7   & 0.96/0.00 & 0.94/0.00 & \textbf{0.99}/0.01 & 0.93/0.00 \\
    Fault 8   & 0.55/0.00 & 0.67/0.00 & 0.00/0.00 & \textbf{0.99}/0.00 \\
    Fault 9   & 0.00/0.00 & 0.00/0.00 & 0.00/0.00 & 0.00/0.00 \\
    Fault 10   & 0.00/0.00 & 0.00/0.00 & 0.95/0.01 & \textbf{0.98}/0.00 \\
    Fault 11   & 0.00/0.00 & 0.00/0.00 & 0.96/0.00 & \textbf{0.99}/0.00 \\
    Fault 12   & 0.00/0.00 & 0.00/0.00 & 0.00/0.00 & \textbf{0.99}/0.00 \\
    Fault 13   & 0.73/0.00 & 0.86/0.00 & 0.95/0.00 & \textbf{0.96}/0.00 \\
    Fault 14   & 0.00/0.00 & 0.00/0.00 & 0.00/0.00 & \textbf{1.00}/0.00 \\
    Fault 15   & 0.00/0.00 & 0.00/0.00 & 0.00/0.00 & 0.00/0.00 \\
    Fault 16   & 0.00/0.00 & 0.00/0.00 & 0.00/0.00 & 0.00/0.00 \\
    Fault 17   & 0.94/0.00 & 0.94/0.00 & \textbf{0.96}/0.00 & 0.91/0.00 \\
    Fault 18   & 0.93/0.00 & 0.00/0.00 & 0.95/0.00 & \textbf{0.96}/0.00 \\
    Fault 19   & 0.79/0.00 & 0.92/0.00 & 0.98/0.00 & \textbf{0.99}/0.00 \\
    Fault 20   & 0.93/0.00 & 0.93/0.00 & 0.94/0.00 & \textbf{0.97}/0.00 \\
    Fault 21   & 0.00/0.00 & 0.00/0.00 & 0.00/0.00 & 0.00/0.00 \\
    Fault 22   & 0.00/0.00 & 0.00/0.00 & 0.00/0.00 & \textbf{0.96}/0.00 \\
    Fault 23   & 0.00/0.00 & 0.00/0.00 & 0.00/0.00 & \textbf{0.96}/0.00 \\
    Fault 24   & 0.00/0.00 & 0.00/0.00 & 0.98/0.00 & \textbf{0.99}/0.00 \\
    Fault 25   & 0.00/0.00 & 0.00/0.00 & \textbf{0.98}/0.00 & 0.95/0.00 \\
    Fault 26   & 0.00/0.00 & 0.94/0.00 & 0.95/0.00 & \textbf{0.98}/0.00 \\
    Fault 27   & 0.00/0.00 & 0.00/0.00 & 0.98/0.00 & \textbf{0.99}/0.00 \\
    Fault 28   & 0.00/0.00 & 0.00/0.00 & 0.00/0.00 & \textbf{0.96}/0.01 \\
    \hline
    \end{tabular}
\end{table}

\end{document}